%% file: 0.main.tex
\documentclass{article}
\PassOptionsToPackage{numbers,sort&compress}{natbib}
\usepackage[preprint]{neurips_2026} 

\usepackage{natbib}
\usepackage{amsmath,amsfonts,amssymb}
\usepackage{algorithm}
\usepackage{algorithmic}
\usepackage{float}
\usepackage{verbatim}
\usepackage{booktabs}
\usepackage{multirow}
\usepackage{graphicx}
\usepackage{soul}
\usepackage{array}

\usepackage{graphicx}
\usepackage{subcaption}

\usepackage[utf8]{inputenc} 
\usepackage[T1]{fontenc}    
\usepackage{hyperref}       
\usepackage{url}            
\usepackage{booktabs}       
\usepackage{amsfonts}       
\usepackage{nicefrac}       
\usepackage{microtype}      
\usepackage{xcolor}         
\usepackage{relsize}
\usepackage{wrapfig}

\usepackage{makecell}
\usepackage[table]{xcolor}

\title{MM++: Post-Hoc Scale-Invariant Multilayer OOD Detection via Top-$K$ Gated Feature Fusion}

\author{
  Rahim Hossain \quad Md Tawheedul Islam Bhuian \quad Md Farhan Shadiq \quad Kyoung-Don Kang \\
  \smallskip
  \small School of Computing \\
  \small State University of New York at Binghamton \\
  \small \texttt{\{rhossain, mislambhuian, mshadiq, kang\}@binghamton.edu}
}

\begin{document}

\maketitle

\begin{abstract}
We introduce MM++ (Multilayer Mahalanobis++), a strictly post-hoc, and scale-invariant framework for Out-of-Distribution (OOD) detection. To address the trade-off between scale invariance and hierarchical expressivity, MM++ constructs a principled joint feature space. It first identifies discriminative intermediate layers by measuring entropy density drops, which mark the boundaries of sharp semantic compression. By fusing these selected layers with the terminal representation, the framework captures latent cross-layer correlations while mitigating early-layer noise. Crucially, a Ledoit–Wolf regularized tied covariance matrix stabilizes this unified space, enabling reliable distance estimation. Requiring no auxiliary OOD data, classifier fine-tuning, or architectural modifications, MM++ delivers robust performance across distinct architectures for both near- and far-OOD detection.
\end{abstract}

\input{1.intro7}
\input{2.related2}
\input{3.method-corrected}

\input{4.eval4}
\input{5.conc}

\bibliographystyle{plainnat}
\bibliography{reference}


\newpage
\appendix
\setcounter{equation}{0}
\renewcommand{\theequation}{A.\arabic{equation}}

\input{7.Appendix}

\newpage

\end{document}

%% file: 1.intro7.tex
\section{Introduction}
\label{sec:intro}

Deep neural networks (DNNs) have achieved remarkable success but often exhibit overconfident predictions on out-of-distribution (OOD) inputs when deployed in open-world environments. This behavior poses significant risks in safety-critical applications such as medical diagnosis and autonomous driving \cite{yang2024generalized, lu2025out}. Consequently, reliable OOD detection remains a fundamental requirement for trustworthy DNN deployment.

A key driver of this overconfidence is a geometric phenomenon known as neural collapse \cite{papyan2020prevalence}. During training, DNNs compress in-distribution (ID) representations into highly structured, low-dimensional class centroids, suppressing intra-class variability. While improving classification accuracy, this terminal compression reduces representational diversity, causing OOD samples to project near ID class structures and yield overconfident predictions.

To mitigate this without retraining, post-hoc methods often leverage intermediate representations \cite{lee2018simple, ren2021simple, sun2022knn, mueller2025mahalanobisPP}. Mahalanobis-based approaches \cite{lee2018simple} model feature activations as class-conditional Gaussian distributions. To address feature scale sensitivity, Mahalanobis++ \cite{mueller2025mahalanobisPP} introduces scale-invariance via unit hypersphere projection prior to distance computation. However, operating solely on the terminally compressed penultimate layer limits its ability to capture mid-level structural cues critical for detecting near-OOD samples.

Multilayer methods like Mahalanobis \cite{lee2018simple} and X-Mahalanobis \cite{xmahalanobis2025} incorporate intermediate layers but inherently treat layer-wise representations as independent marginals, relying on the additive fusion of individual scores. This mathematical simplification discards cross-layer conditional dependencies. Consequently, they are less capable to detect hierarchical inconsistencies, where an anomaly mimics ID features at individual layers but violates the expected evolutionary trajectory between them.

Furthermore, aggregating these independent scores requires regressing fusion weights using proxy OOD validation sets or classifier fine-tuning. This violates the strictly post-hoc assumption, biasing the detector toward the specific geometry of auxiliary anomalies.

A naive integration of Mahalanobis++'s scale-invariance with X-Mahalanobis's multi-layer extraction remains limited if the underlying issue of marginal score aggregation persists. Robust multi-layer OOD detection requires modeling the unified joint distribution of the feature hierarchy, driven by intrinsic ID geometry, instead of additive score fusion.

To address this, we propose \textbf{MM++} (Multilayer Mahalanobis++), a \textit{strictly post-hoc}, \textit{scale-invariant} and  framework. MM++ eschews ad hoc layer weighting and score addition. Instead, it extends hyperspherical normalization to intermediate representations and introduces a Top-$K$ information gating mechanism to systematically select the most informative layers.

Specifically, we quantify layer-wise informativeness using covariance entropy estimated via Ledoit--Wolf shrinkage \cite{ledoit2004well}. We derive an entropy density drop ($\Delta_l$) to identify layers undergoing the sharpest semantic compression. The penultimate layer is included as an anchor, while the top $K-1$ intermediate layers are selected based on $\Delta_l$, ensuring focus on representations with the most discriminative structural transitions.

\begin{figure}[htbp]
    \centering
    \includegraphics[width=\textwidth]{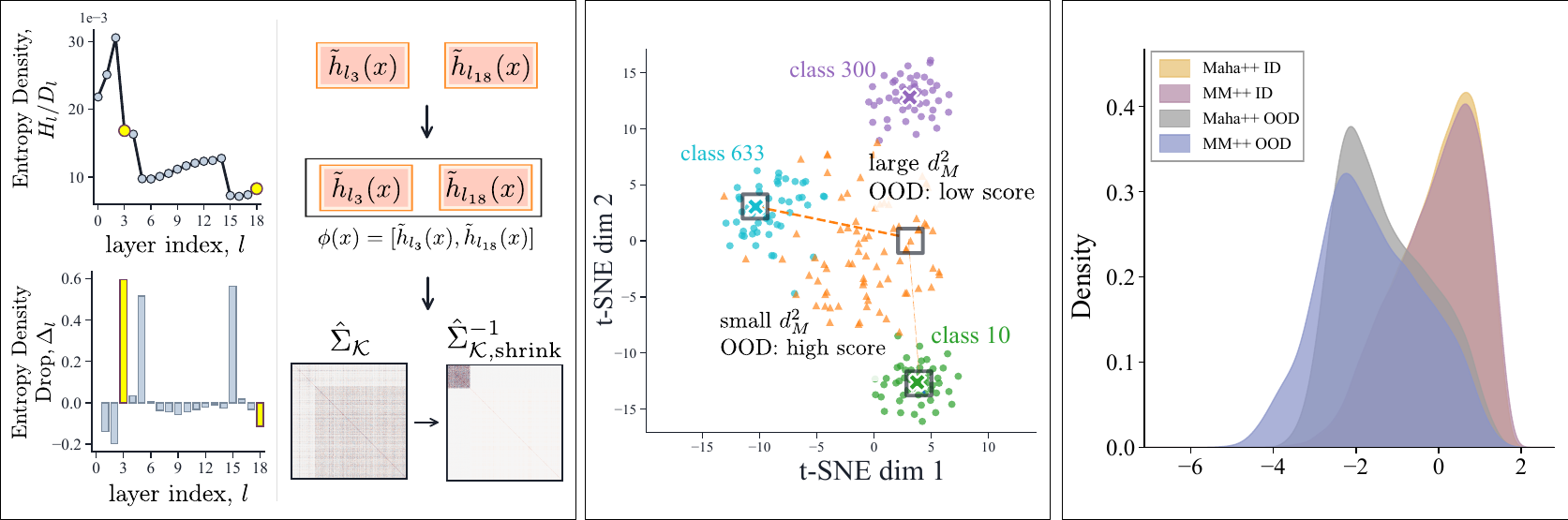}
    \caption{\textbf{Overview of the MM++ Framework.} Post-hoc, scale-invariant multi-layer OOD detection via top-$K$ gated feature fusion (illustrated with $K = 2$, ConvNeXt-T on ImageNet-LT as ID, and ImageNet-C as OOD). \textbf{Left (Pipeline):} MM++ first identifies top-$K$ layers by leveraging entropy density drops to capture maximum cross-layer compression, anchoring the penultimate layer. These intermediate and terminal features are concatenated into a unified representation $\phi(x)$, stabilized by a well-conditioned, shrunken tied covariance matrix ($\hat{\Sigma}_{\mathcal{K},\text{shrink}}$). \textbf{Center (Feature Space):} In this unified space, ID samples tightly cluster around established class centroids (yielding small Mahalanobis++ distances, $d_M^2$), while OOD samples are explicitly pushed to the periphery. \textbf{Right (Distributions):} Consequently, MM++ significantly tightens the ID density and isolates the OOD distribution. By pushing OOD samples into a negative tail, it achieves substantially higher separability and reduces ID-OOD overlap compared to state-of-the-art baselines.}
    \label{fig:mmpp_teaser}
\end{figure}

The selected $\ell_2$-normalized features are concatenated into a unified representation, on which a single joint Mahalanobis++ distance is computed. By modeling cross-layer dependencies through a shared precision matrix—estimated via Ledoit–Wolf shrinkage—MM++ effectively penalizes samples that exhibit anomalous hierarchical trajectories. Conceptually, this fusion mechanism generalizes across heterogeneous architectures, as it operates on normalized intermediate features without requiring explicit layer-wise weighting or architecture-specific design choices (see Figure~\ref{fig:mmpp_teaser}).

\textbf{Contributions.}

\textbf{MM++: A Unified Multilayer Framework.}
We propose a strictly post-hoc framework that shifts the multi-layer paradigm from additive score aggregation to joint feature-space modeling. MM++ integrates (i) a layer selection mechanism using covariance entropy and entropy density drops; (ii) architecture-independent concatenation of $\ell_2$-normalized features anchored at the penultimate layer; and (iii) a single joint Mahalanobis++ estimator with Ledoit--Wolf-regularized precision to explicitly capture cross-layer dependencies. It requires no proxy OOD data or fine-tuning, and introduces only a single hyperparameter ($K$).

\textbf{Empirical Validation.} We evaluate MM++ across global attention (ViTs), hierarchical attention (Swin), and convolutional (ConvNeXt) backbones. Unlike specialized methods such as X-Mahalanobis \cite{xmahalanobis2025}, which are tailored to Transformers and balanced data, MM++ offers a more universal solution. It delivers consistently robust performance on ViTs, while extending state-of-the-art multilayer OOD detection to convolutional and hierarchical paradigms. Furthermore, in the long-tailed ImageNet-LT regime, MM++ consistently outperforms baselines on challenging near-OOD benchmarks (ImageNet-V2, -C, -ES, -R), demonstrating higher resilience to both architectural variance and class imbalance.

%% file: 2.related2.tex
\section{Related Work}
\label{sec:related_work}

OOD detection has evolved from simple output-based heuristics to approaches that analyze the geometric structure of deep representations \cite{yang2024generalized, lu2025out}. We categorize prior work into five directions, focusing on post-hoc methods most relevant to our setting. We exclude training-based approaches, such as \cite{hendrycks2019using, sehwag2021ssd, du2022towards, hendrycks2022pixmix, ming2022exploit, tao2023non, lu2024learning, regmi2024t2fnorm, Haas2023ExploringSH}, as they require additional training and are orthogonal to the strictly post-hoc regime considered in this work.

\textbf{Output- and Logit-Based Methods.}
Early post-hoc methods rely on final network outputs. \cite{hendrycks2017baseline} showed that maximum softmax probability is typically lower for OOD samples. ODIN \cite{liang2018enhancing} improves separability via temperature scaling and input perturbation, while energy-based scores \cite{liu2020energy} reinterpret logits through the lens of energy-based models. Despite their efficiency, these approaches are constrained by their reliance on the low-dimensional outputs of the final linear classification layer, which discard rich intermediate representations and limit access to geometric structures, such as those associated with neural collapse, that can be informative for robust OOD detection.

\textbf{Feature-Based and Geometric Methods.}
To overcome this limitation, feature-based approaches operate in the pre-logit space. A seminal method is the Mahalanobis distance \cite{lee2018simple}, which models features as class-conditional Gaussians with shared covariance and measures distances to class centers. Extensions include non-parametric methods such as $k$-nearest neighbors \cite{sun2022knn} and relative Mahalanobis variants \cite{ren2021simple}. These approaches have also been applied beyond vision, including language modeling \cite{hofmann2026ap} and medical imaging \cite{anthony2023use, wei2023fine}. However, their effectiveness depends on simplifying distributional assumptions, such as Gaussianity, that are often misaligned with the complex, anisotropic feature geometries observed in modern architectures \cite{bitterwolf2023ninco, mueller2025mahalanobisPP}.

\textbf{Scale Invariance and Feature Rectification.}
Modern architectures often produce feature embeddings with substantial variation in magnitude, which can distort distance-based metrics. ReAct \cite{sun2021react} addresses this by truncating extreme activations. More recently, Mahalanobis++ \cite{mueller2025mahalanobisPP} introduces scale invariance by projecting features onto a unit hypersphere prior to distance computation. This improves robustness by reducing norm-induced bias, but remains restricted to the penultimate layer.

\textbf{Multilayer and Hierarchical Detection.}
Deep networks encode information hierarchically, from low-level features to high-level semantics. Motivated by this, multilayer approaches aim to detect OOD signals across depth \cite{lee2018simple, lin2021mood, sun2022knn}. 
In particular, X-Mahalanobis \cite{xmahalanobis2025} attempts to capture distributional shifts by densely aggregating layer-wise distance scores. While conceptually unsupervised, it relies on parameter-efficient fine-tuning of a task-specific linear head, rendering it not strictly post-hoc in practice. Moreover, its dense aggregation scheme via variance-based weighting may assign non-negligible weights to early layers, which may allow high-variance spatial noise to propagate and interfere with the deeper semantic representations essential for reliable OOD detection.


\textbf{Covariance Entropy and Sparsification.}
Recent theory highlights the information bottleneck principle \cite{tishby2015deep} and neural collapse \cite{papyan2020prevalence}, where deep features concentrate into low-dimensional subspaces as semantic information becomes increasingly aligned and nuisance variation is progressively suppressed. The structure of these representations can be quantified via the entropy of the covariance eigenspectrum \cite{harun2025controlling, janiak2025geometry, neco2024}. Layers exhibiting sharp decreases in this entropy, indicating rapid semantic compression, may correspond to critical transition points in the hierarchical representation.
However, standard soft-weighting schemes based on raw entropy values do not induce sparsity across layers and therefore retain contributions from less compressed representations. This may introduce residual early-layer variability into the aggregated representation \cite{martins2016softmax, shwartz2017opening}. More fundamentally, such continuous weighting obscures discrete structural changes in representation geometry, making it difficult to localize compression boundaries.
Motivated by this observation, we instead seek a discrete criterion that identifies abrupt changes in representational complexity. Specifically, we define entropy density drops to detect sharp transitions in covariance structure, and use these as indicators for selecting a small subset of semantically compressed layers. This enables selective Top-$K$ gating that emphasizes post-collapse representations while suppressing high-variance early-layer signals.

\textbf{Connection to Our Work.} MM++ builds upon these directions to reconcile the inherent trade-off between hierarchical expressivity and terminal compression in a strictly post-hoc setting. It extends the scale-invariant $\ell_2$-normalization of Mahalanobis++ to a multi-layer framework, replacing heuristic aggregation—such as the variance-based weighting utilized by X-Mahalanobis \cite{xmahalanobis2025}—with a principled Top-$K$ information gating mechanism. Specifically, we analyze the ID feature spectrum via entropy density drops to pinpoint the boundaries of sharp semantic compression. We anchor the terminal layer, select the $K-1$ most informative intermediate stages, fuse features into a unified representation space via architecture-independent concatenation. This joint space is stabilized by a Ledoit–Wolf regularized tied covariance matrix, allowing a single Mahalanobis++ distance to capture vital cross-layer geometric interactions while suppressing high-variance noise from early layers or under-represented classes. Consequently, MM++ achieves robust near- and far-OOD detection across heterogeneous architectures and long-tailed regimes without requiring proxy OOD data or classifier fine-tuning.


%% file: 3.method-corrected.tex
\section{Method}
\label{sec:method}

\subsection{Preliminaries}

Let $\mathcal{D}_{\text{train}} = \{(x_i, y_i)\}_{i=1}^{N}$ be the
in-distribution training set, where $y_i \in \{1,\ldots,C\}$ and $N_c$
denotes the number of samples in class $c$.  A pre-trained deep network
$\mathcal{F} = g \circ h$ decomposes into a feature extractor $h$ and a
classifier head $g$.
We denote the feature extractor $h$ as an $L$-layer network.
Thus, layer $L$ represents the terminal feature space of $h$, which serves as
the penultimate layer of the overall network $\mathcal{F}$.
At inference, we extract intermediate feature maps
$h_l(x) \in \mathbb{R}^{D_l}$ from $L$ designated layers
$l = 1, \ldots, L$.

To eliminate feature-magnitude variations across layers and architectures,
we $\ell_2$-normalize every representation onto the unit hypersphere:
\begin{equation}
    \tilde{h}_l(x) = {h_l(x)}/ \|h_l(x)\|_2.
    \label{eq:l2norm}
\end{equation}
For each layer $l \in \{1, \ldots, L\}$, we estimate class-conditional means and a tied
covariance matrix from the normalized training features:
\begin{equation}
    \hat{\mu}_{c,l} = \frac{1}{N_c} \sum_{i:\,y_i=c} \tilde{h}_l(x_i), \; \; \; \; \;
    \hat{\Sigma}_l = \frac{1}{N} \sum_{c=1}^{C} \sum_{i:\,y_i=c}
        \bigl(\tilde{h}_l(x_i) - \hat{\mu}_{c,l}\bigr)
        \bigl(\tilde{h}_l(x_i) - \hat{\mu}_{c,l}\bigr)^{\!\top}.
    \label{eq:tied_cov}
\end{equation}
Because $\hat{\Sigma}_l$ can be rank-deficient or ill-conditioned when $D_l >> N_c$,  we apply Ledoit--Wolf shrinkage \cite{ledoit2004well} to obtain a \textit{well-conditioned} 
covariance matrix:
\begin{equation}
    \hat{\Sigma}_{l,\text{shrink}} =
        (1 - \gamma)\,\hat{\Sigma}_l
        + \gamma\,\frac{\operatorname{Tr}(\hat{\Sigma}_l)}{D_l}\,\mathbf{I}.
    \label{eq:shrinkage}
\end{equation}
where $\gamma \in [0,1]$ is \textit{not a hyperparameter but derived analytically} by minimizing the expected
mean squared error (Frobenius norm) only using ID
data~\cite{ledoit2004well}. Subsequently, the corresponding precision matrix, $\hat{\Sigma}^{-1}_{l,\text{shrink}}$, is computed.


\subsection{Intra-Class Covariance Entropy}


We characterize the representational richness of each layer $l$ in the encoder through the Shannon entropy of the normalized, well-conditioned eigenvalue spectrum of $\hat{\Sigma}_{l, \text{shrink}}$. Let $\{\lambda_i^{(l)}\}_{i=1}^{D_l}$ be the eigenvalues of the shrunk covariance matrix $\hat{\Sigma}_{l,\text{shrink}}$. Because $\hat{\Sigma}_{l,\text{shrink}}$ is positive semi-definite, all $\lambda_i^{(l)} \ge 0$. To ensure numerical stability, we set $\lambda_i^{(l)} = \max(\lambda_i^{(l)}, \epsilon)$ before normalization, where $\epsilon$ is a small positive constant (e.g., $10^{-8}$). Given that, we define the normalized eigenspectrum and the intra-class covariance entropy as:
\begin{equation}
    \bar{\lambda}_i^{(l)} = \lambda_i^{(l)} / \sum_{j=1}^{D_l} \lambda_j^{(l)}, \quad
    H_l = -\sum_{i=1}^{D_l}\bar{\lambda}_i^{(l)}\ln\bar{\lambda}_i^{(l)}.
    \label{eq:entropy}
\end{equation}


\subsection{Layer Selection via Entropy Density Drops}


We define the entropy density of layer $l$---representing the Shannon entropy per feature dimension---as $\rho_l = \frac{H_l}{D_l}$. This formulation yields a scale-invariant metric of representational richness that naturally accounts for varying layer widths $D_l$. To quantify the stage-wise feature compression across the network, we define the \textit{entropy density drop} between consecutive layers as the straightforward difference in their entropy densities:
\begin{equation}
\Delta_l = \rho_{l-1} - \rho_l = \frac{H_{l-1}}{D_{l-1}} - \frac{H_l}{D_l},
\quad l = 2, \ldots, L.
\label{eq:delta}
\end{equation}

A large positive $\Delta_l$ signals a sharp drop in relative intrinsic dimensionality at layer $l$: the network undergoes strong semantic compression at that transition. Layers at such boundaries carry the most
discriminative geometric structure for OOD detection.

Motivated by the neural collapse phenomenon~\citep{papyan2020prevalence},
which indicates that the penultimate layer maximally encodes
class-discriminative geometry, we always include the penultimate layer $L$ in the
active set.  We then select $K{-}1$ additional layers by their entropy density drops:
\begin{equation}
    \mathcal{K} = \{L\} \;\cup\;
        \operatorname*{arg\,top\text{-}(\textit{K}-1)}_{l \in \{2,\ldots,{L-1}\}}
        \Delta_l \text{ where } \Delta_l > 0
    \label{eq:topk_sel}
\end{equation}

If $\Delta_l < 0$, the entropy density increases from layer $l-1$ to layer $l$. 
This can happen in structural expansions (e.g., channel upsampling in hierarchical backbones) or attention-based mixing.



\subsection{Feature Fusion and OOD Scoring}
\label{sec:fusion}

Given the selected layer set $\mathcal{K} = \{l_1, \ldots, l_K\}$, we combine the per-layer representations into a single vector $\boldsymbol{\phi}(x)$ on which a class-conditional Mahalanobis detector is applied.

\textbf{Cross-Layer Concatenated Fusion.}
We propose retaining the full, uncompressed information from every
selected layer by concatenating the $\ell_2$-normalized feature vectors:
\begin{equation}
    \boldsymbol{\phi}(x_i)
        = \bigl[\,\tilde{h}_{l_1}(x_i);
        \;\ldots;\;
                  \tilde{h}_{l_K}(x_i)\,\bigr]
        \;\in\; \mathbb{R}^{\sum_{l \in \mathcal{K}} D_l}.
    \label{eq:fuse_concat}
\end{equation}
This representation is lossless for each layer $l \in \mathcal{K}$: it encodes the per-layer features without any mixing. Moreover, it is \textit{architecture-independent:}
for homogeneous architectures (e.g.\ ViT with $D_l = D$), the joint
dimension is $D_{\mathcal{K}} = \sum_{l \in \mathcal{K}} D_l = K \times D$. 
For heterogeneous architectures
(e.g., Swin-T or ConvNeXt with varying $D_l$), 
$D_{\mathcal{K}} = \sum_{l \in \mathcal{K}} D_l$.
Crucially, \textit{no explicit aggregation weights are required}: the scalar
contribution of each layer is learned implicitly by the joint precision
matrix described below. Thus, MM++ introduces only one hyperparameter ($K$).

\textbf{Joint Precision Matrix Estimation.}
We compute fused class means and the tied covariance matrix for the selected $K$ layers:
\begin{equation}
    \hat{\boldsymbol{\mu}}_c^{\mathcal{K}}
        = \frac{1}{N_c} \sum_{i:\,y_i=c} \boldsymbol{\phi}(x_i), \quad
    \hat{\Sigma}_\mathcal{K} = \frac{1}{N} \sum_{c=1}^{C} \sum_{i:\,y_i=c}
        \bigl(\boldsymbol{\phi}(x_i) - \hat{\boldsymbol{\mu}}_{y_i}^{\mathcal{K}}\bigr)
        \bigl(\boldsymbol{\phi}(x_i) - \hat{\boldsymbol{\mu}}_{y_i}^{\mathcal{K}}\bigr)^{\!\top}.    
    \label{eq:fused_mean}
\end{equation}
We then apply Ledoit--Wolf shrinkage to the centered fused representations to obtain a well-conditioned tied covariance matrix $\hat{\Sigma}_{\mathcal{K},\text{shrink}}$:
\begin{equation}
    \hat{\Sigma}_{\mathcal{K},\text{shrink}} = (1 - \gamma) \hat{\Sigma}_{\mathcal{K}} + \gamma\,\left(\frac{\operatorname{Tr}(\hat{\Sigma}_{\mathcal{K}})}{D_{\mathcal{K}}}\right)\mathbf{I}   
      \;\in\; \mathbb{R}^{D_{\mathcal{K}} \times D_{\mathcal{K}}},
    \label{eq:joint_prec}
\end{equation}
where $\mathbf{I}$ is an identity matrix with dimension $D_{\mathcal{K}} \times D_{\mathcal{K}}$.

Let $\hat{\Sigma}_{\mathcal{K}}^{-1}
=\hat{\Sigma}_{\mathcal{K},\text{shrink}}^{-1}$ denote the shrunk precision matrix in the fused representation space.
The diagonal blocks $\hat{\Sigma}_{\mathcal{K},ll}^{-1}$ capture
intra-layer class-conditional covariance at each selected layer $l$,
recovering the behavior of single-layer Mahalanobis++ detectors.
The off-diagonal blocks $\hat{\Sigma}_{\mathcal{K},ll'}^{-1}$ ($l \neq l'$)
encode \textit{cross-layer covariance}, capturing how representations at
layers $l$ and $l'$ co-vary within a class.
Such interactions cannot be modeled by additive fusion, which reduces the
precision to a single $D \times D$ matrix and conflates intra- and
cross-layer structure.
Ledoit--Wolf shrinkage ensures a well-conditioned estimate even when
$\sum_{l \in \mathcal{K}} D_l \gg N_c$ \cite{ledoit2004well}, with the
shrinkage coefficient computed analytically from ID training data.

\textbf{OOD Score of MM++.}
The final OOD confidence score for a test input $x$ is the negative
minimum Mahalanobis++ distance over all classes in the
joint feature space:
\begin{equation}
    \mathcal{S}_{\text{MM++}}(x) = -\min_{c \in \{1,\ldots,C\}}\;
        \bigl(\boldsymbol{\phi}(x) - \hat{\boldsymbol{\mu}}_c^{\mathcal{K}}\bigr)^{\!\top}
        \hat{\Sigma}_{\mathcal{K}}^{-1}
        \bigl(\boldsymbol{\phi}(x) - \hat{\boldsymbol{\mu}}_c^{\mathcal{K}}\bigr).
    \label{eq:score}
\end{equation}

Higher values of $\mathcal{S}_{\text{MM++}}(x)$ indicate that $x$ is closer to some
in-distribution class cluster and is therefore more likely in-distribution.
A test sample may appear plausible under the single-layer distributions at
$l_1$ and $l_2$ individually, yet be flagged by
$\hat{\Sigma}_{\mathcal{K}}^{-1}$ because the joint configuration
$(\tilde{h}_{l_1}(x), \tilde{h}_{l_2}(x))$ is inconsistent with any
training class---a \textit{failure mode} that additive fusion cannot detect.
This method requires neither labeled OOD data nor additional model training/fine-tuning.  Instead, it relies solely on closed-form Ledoit--Wolf covariance estimates computed from ID data.

In MM++, calibration that estimates $\hat{\boldsymbol{\mu}}_c^{\mathcal{K}}$ and $\hat{\Sigma}_{\mathcal{K}}^{-1}$ in the unified space is performed once offline. 
Using them, MM++ performs
online OOD detection for test input $x$ (Eq~\ref{eq:score}). 
The pseudocode for MM++ is presented in Algorithm~\ref{alg:mm_plus_plus}, and a theoretical justification is provided in Appendix~\ref{app:justify}.

\begin{algorithm}[H]
\caption{MM++: Offline Calibration and Online Inference}
\label{alg:mm_plus_plus}
\small
\begin{algorithmic}[1]
\REQUIRE ID dataset $\mathcal{D}_{\text{ID}}$, pre-trained encoder $h$ with $L$ layers, fusion size $K$, test sample $x$
\ENSURE OOD score $\mathcal{S}_{\text{MM++}}(x)$

\STATE \textbf{\textsc{Phase I: One-Time Offline Calibration}} 
\FOR{$l \in \{1,\dots,L\}$}
    \STATE Extract and $\ell_2$-normalize features $\tilde{h}_l(x_i)$ for all training samples $x_i \in \mathcal{D}_{\text{ID}}$
    \STATE Estimate within-class covariance $\hat{\Sigma}_l$ and apply Ledoit--Wolf shrinkage
    \STATE Compute covariance entropy $H_l$
\ENDFOR
\STATE Compute entropy density drops $\Delta_l$ for $l \ge 2$ and select layers: $\mathcal{K} = \{L\} \cup \arg\text{top}_{K-1}(\{\Delta_l\}_{l=2}^{L-1})$

\FOR{each sample $x_i \in \mathcal{D}_{\text{ID}}$}
    \STATE Fuse by concatenation: $\boldsymbol{\phi}(x_i) = \bigl[\,\tilde{h}_{l_1}(x_i); \;\ldots;\; \tilde{h}_{l_K}(x_i)\,\bigr]$
\ENDFOR
\FOR{class $c \in \{1,\dots,C\}$}
    \STATE Compute fused class mean: $\hat{\boldsymbol{\mu}}_c^{\mathcal{K}} = \frac{1}{N_c} \sum_{i:\,y_i=c} \boldsymbol{\phi}(x_i)$
\ENDFOR
\STATE Compute tied covariance $\hat{\Sigma}_{\mathcal{K}}$ and apply Ledoit--Wolf shrinkage to obtain precision matrix $\hat{\Sigma}_{\mathcal{K}}^{-1}$

\STATE \textbf{\textsc{Phase II: Test-Time Inference}}
\STATE For test input $x$, extract and $\ell_2$-normalize features $\tilde{h}_l(x)$ for selected layers $l \in \mathcal{K}$
\STATE Construct fused representation: $\boldsymbol{\phi}(x) = \bigl[\,\tilde{h}_{l_1}(x); \;\ldots;\; \tilde{h}_{l_K}(x)\,\bigr]$ 
\STATE \textbf{return} $\mathcal{S}_{\text{MM++}}(x) = -\min_{c}\; \bigl(\boldsymbol{\phi}(x) - \hat{\boldsymbol{\mu}}_c^{\mathcal{K}}\bigr)^{\!\top} \hat{\Sigma}_{\mathcal{K}}^{-1} \bigl(\boldsymbol{\phi}(x) - \hat{\boldsymbol{\mu}}_c^{\mathcal{K}}\bigr)$
\end{algorithmic}
\end{algorithm}

%% file: 4.eval4.tex
\section{Evaluation}
\label{sec:eval}

\subsection{Experimental Setup}

\textbf{In-Distribution Dataset ($\mathcal{D}_{\text{ID}}$):} We evaluate MM++ on ImageNet-1K \cite{russakovsky2015imagenet} and its long-tailed counterpart, ImageNet-LT \cite{liu2019large}, as primary ID benchmarks. This enables a rigorous assessment of the framework’s stability under both balanced and skewed class distributions. The specific ID dataset is indicated in each respective results discussion.

\textbf{Out-of-Distribution Datasets:} We evaluate challenging distribution shifts covering near-OOD datasets with high semantic similarity (ImageNet-V2 \cite{Rechtetal2019}, ImageNet-C \cite{hendrycks2018benchmarking}, ImageNet-R \cite{Hendrycksetal2021}, ImageNet-ES \cite{Linetal2021}) and far-OOD datasets exhibiting strong domain shifts (ImageNet-O \cite{hendrycks2021natural}, OpenImage-O \cite{yang2022openood}, NINCO \cite{bitterwolf2023ninco}, Places365 \cite{zhou2017places}, Textures \cite{cimpoi2014describing}, iNaturalist \cite{VanHornetal2018}, SUN \cite{Xiaoetal2013}).

\textbf{Metrics.} We report AUROC and FPR95. The threshold $\tau$ is set such that $P(\mathcal{S}(x) \ge \tau \mid x \in \mathcal{D}_{\text{ID}})=0.95$. An input $x$ is classified as ID if $\mathcal{S}(x) \ge \tau$, and as OOD otherwise.

\textbf{Backbones.} Main results are reported using an ImageNet-21K pretrained ViT-B/16~\cite{dosovitskiy2020image}. Appendix~\ref{app:more-results} provides additional results demonstrating cross-architectural applicability across diverse design paradigms, including Swin-T~\cite{Liu_2021_ICCV} (hierarchical attention), ConvNeXt-T~\cite{liu2022convnet} (modernized convolutions), and EVA02-S14~\cite{fang2024eva} (a large-scale pretrained vision transformer with improved representation capacity).

\textbf{Baselines.} We compare MM++ against established post-hoc methods: MSP \cite{hendrycks2017baseline}, ODIN \cite{liang2018enhancing}, Energy \cite{liu2020energy}, ReAct \cite{sun2021react}, KNN \cite{sun2022knn}, Mahalanobis \cite{lee2018simple}, Relative Mahalanobis \cite{ren2021simple}, Mahalanobis++ \cite{mueller2025mahalanobisPP}, Relative Mahalanobis++ (an $\ell_2$-normalized variant for evaluation), and X-Mahalanobis \cite{xmahalanobis2025} (reviewed in Appendix~\ref{app:methods}).

We use public implementations with frozen parameters and consistent datasets. Only X-Mahalanobis fine-tunes the classifier \cite{xmahalanobis2025}. For MM++, we use $K=2$ (the anchored penultimate layer and one information-gated intermediate layer). A sensitivity analysis for $K$ is presented in Section~\ref{sec:ablations}.

\subsection{Comparison to State-of-the-Art}
\label{sec:sota}

\textbf{Results using ImageNet-1K as ID.}
As shown in Table~\ref{tab:main_results}, specialized methods such as X-Mahalanobis—which fine-tunes on the ID training set—achieve the highest overall performance on the balanced ImageNet-1K benchmark, proving effective at establishing broad decision boundaries for far-OOD outliers. Nevertheless, MM++ offers a versatile, strictly post-hoc alternative, achieving competitive performance and the best AUROC on Texture and SUN without requiring auxiliary OOD data, extensive hyperparameter calibration, or fine-tuning.

\textbf{Results using ImageNet-LT as ID.}
Table~\ref{tab:imagenetlt_results} evaluates the challenging long-tailed ID regime. 
Here, MM++ demonstrates superior robustness, yielding the highest average AUROC (83.91\%). While X-Mahalanobis’s fine-tuning grants an advantage on pure semantic shifts (NINCO, OpenImage-O), MM++ excels under severe near-OOD shifts (ImageNet-ES, -R, -V2). By integrating structural correlations from intermediate blocks with the anchored penultimate layer, MM++ effectively handles subtle semantic deviations that terminal-layer representations or additive multi-layer fusions, which aggregate individual layer-wise scores, are likely to overlook in imbalanced regimes.

\colorlet{highlightcolor}{blue!10} 

\begin{table*}[htb]
\centering
\caption{OOD detection performance using ViT-B/16 with ImageNet-1K as ID.
($\dagger$ denotes original paper results. $\uparrow$: higher is better, $\downarrow$: lower is better; \textbf{best} and \underline{second-best}; values are in \%.)}
\label{tab:main_results}
\small
\setlength{\tabcolsep}{2pt}
\renewcommand{\arraystretch}{1.6}
\resizebox{\textwidth}{!}{%
\begin{tabular}{>{\Large}l >{\Large}c>{\Large}c >{\Large}c>{\Large}c >{\Large}c>{\Large}c >{\Large}c>{\Large}c >{\Large}c>{\Large}c >{\Large}c>{\Large}c >{\Large}c>{\Large}c}
\toprule
\multirow{2}{*}{\textbf{Method}} &
\multicolumn{2}{c}{\large \textbf{ImageNet-O}} &
\multicolumn{2}{c}{\large \textbf{Texture}} &
\multicolumn{2}{c}{\large \textbf{Places365}} &
\multicolumn{2}{c}{\large \textbf{iNaturalist}} &
\multicolumn{2}{c}{\large \textbf{SUN}} &
\multicolumn{2}{c}{\large \textbf{Average}} \\ 
& {\large AUROC $\uparrow$} & {\large FPR95 $\downarrow$}
& {\large AUROC $\uparrow$} & {\large FPR95 $\downarrow$}
& {\large AUROC $\uparrow$} & {\large FPR95 $\downarrow$}
& {\large AUROC $\uparrow$} & {\large FPR95 $\downarrow$}
& {\large AUROC $\uparrow$} & {\large FPR95 $\downarrow$}
& {\large AUROC $\uparrow$} & {\large FPR95 $\downarrow$}\\
\midrule
MSP     & 73.63 & 75.30 & 82.72 & 58.39 & 80.44 & 65.90 & 92.57 & 34.35 & 84.75 & 55.98 & 82.82 & 57.98 \\
ODIN    & 65.42 & 75.10 & 72.78 & 65.50 & 69.69 & 75.09 & 85.80 & 41.45 & 77.51 & 60.73 & 74.24 & 63.57 \\
Energy  & 75.97 & 60.00 & 83.53 & 47.07 & 73.16 & 68.95 & 91.59 & 29.73 & 83.25 & 49.06 & 81.50 & 50.96 \\
ReAct   & 79.20 & 56.95 & 85.53 & 44.45 & 78.37 & 61.87 & 95.54 & 18.75 & 84.81 & 47.59 & 84.69 & 45.52 \\
KNN     & 84.98 & 67.90 & 89.56 & 42.93 & 85.15 & 61.54 & 96.57 & 16.20 & 87.48 & 55.26 & 88.75 & 48.77 \\


\midrule
Maha    & 83.65 & 75.70 & 85.43 & 64.57 & 84.57 & 64.58 & 96.98 & 13.14 & 85.97 & 64.26 & 87.32 & 56.45 \\
rMaha   & 84.42 & 70.40 & 86.89 & 58.79 & 85.28 & 63.09 & 97.61 & 10.62 & 87.41 & 59.11 & 88.32 & 52.40 \\
Maha++  & 86.42 & 63.70 & 89.26 & 48.26 & 85.75 & 64.11 & \underline{98.76} & \underline{5.15} & 88.90 & 53.47 & 89.82 & 46.94 \\
rMaha++ & 86.45 & 63.65 & 89.31 & 48.26 & 85.77 & 64.32 & \underline{98.77} & 5.16 & \underline{88.93} & 53.63 & 89.85 & 47.00 \\
\midrule
$\text{X-Maha}^{\dagger}$ & \textbf{93.76} & \textbf{29.80} & \underline{96.65} & \textbf{11.70} & \textbf{92.04} & \textbf{37.78} & \textbf{99.40} & \textbf{2.26} & \textbf{89.64} & \textbf{46.00} & \textbf{94.30} & \textbf{25.51} \\


MM++ (Ours)    & \underline{88.76} & \underline{54.45} & \textbf{97.00} & \underline{14.52} & \underline{87.05} & \underline{55.58} & 98.43 & 5.71 & \textbf{89.64} & \underline{49.66} & \underline{92.18} & \underline{35.98} \\
\bottomrule
\end{tabular}%
}
\end{table*}

\begin{table*}[htb]
\centering
\caption{OOD detection performance using ViT-B/16 with ImageNet-LT as ID.}
\label{tab:imagenetlt_results}
\small
\setlength{\tabcolsep}{2pt}
\renewcommand{\arraystretch}{1.9}
\resizebox{\textwidth}{!}{%
\begin{tabular}{>{\LARGE}l >{\LARGE}c>{\LARGE}c >{\LARGE}c>{\LARGE}c >{\LARGE}c>{\LARGE}c >{\LARGE}c>{\LARGE}c >{\LARGE}c>{\LARGE}c >{\LARGE}c>{\LARGE}c >{\LARGE}c>{\LARGE}c}
\toprule
\multirow{2}{*}{\textbf{Method}} &
\multicolumn{2}{c}{\Large \textbf{NINCO}} &
\multicolumn{2}{c}{\Large \textbf{OpenImage-O}} &
\multicolumn{2}{c}{\Large \textbf{ImageNet-C}} &
\multicolumn{2}{c}{\Large \textbf{ImageNet-ES}} &
\multicolumn{2}{c}{\Large \textbf{ImageNet-R}} &
\multicolumn{2}{c}{\Large \textbf{ImageNet-V2}} &
\multicolumn{2}{c}{\Large \textbf{Average}} \\
& {\large AUROC $\uparrow$} & {\large FPR95 $\downarrow$}
& {\large AUROC $\uparrow$} & {\large FPR95 $\downarrow$}
& {\large AUROC $\uparrow$} & {\large FPR95 $\downarrow$}
& {\large AUROC $\uparrow$} & {\large FPR95 $\downarrow$}
& {\large AUROC $\uparrow$} & {\large FPR95 $\downarrow$}
& {\large AUROC $\uparrow$} & {\large FPR95 $\downarrow$}
& {\large AUROC $\uparrow$} & {\large FPR95 $\downarrow$}\\
\midrule
MSP  & 83.28 & 61.85 & 89.23 & 44.89 & 71.21 & 73.94 & 69.35 & 70.01 & 77.63 & 66.23 & 58.73 & 90.14 & 74.24 & 67.84 \\
ODIN  & 74.28 & 64.69 & 81.93 & 48.20 & 64.61 & 79.35 & 63.22 & 74.31 & 67.90 & 74.49 & 54.82 & 92.10 & 67.79 & 72.52 \\
Energy  & 83.23 & 51.22 & 89.16 & 33.76 & 72.34 & 66.65 & 67.54 & 70.31 & 72.87 & 67.04 & 57.29 & 90.85 & 73.74 & 63.31 \\
ReAct  & 85.62 & 48.58 & 92.76 & 26.74 & 73.55 & 65.85 & 69.57 & 67.84 & 78.19 & 60.11 & 58.65 & 89.91 & 76.39 & 59.84 \\
KNN  & 40.44 & 96.00 & 50.34 & 95.06 & 57.73 & 88.45 & 55.88 & 91.43 & 65.61 & 84.17 & 49.80 & 94.34 & 53.3 & 91.91 \\
\midrule
Maha  & 85.77 & 70.01 & 93.54 & 38.54 & 70.94 & 85.00 & 69.53 & 84.53 & 86.23 & 51.85 & 58.29 & 89.76 & 77.38 & 69.95 \\
rMaha  & 88.51 & 57.85 & 94.62 & 29.87 & 71.73 & 80.32 & 71.30 & 80.22 & 85.02 & 52.00 & 59.18 & 89.60 & 78.39 & 64.98 \\
Maha++  & 90.37 & 47.30 & 95.81 & 24.55 & 74.86 & 73.05 & 73.51 & 73.09 & 87.22 & 46.92 & 59.52 & 89.37 & 80.22 & 59.71 \\
rMaha++  & 90.48 & 44.33 & 95.47 & 25.24 & 72.49 & 74.97 & 71.90 & 72.93 & 85.93 & 49.18 & 59.08 & \underline{89.34} & 79.22 & 59.00 \\
\midrule
$\text{X-Maha}^{\dagger}$ & \textbf{94.98} & \textbf{26.74} & \textbf{98.21} & \textbf{9.72} & \underline{83.96} & \textbf{53.57} & \underline{76.78} & \underline{63.45} & \underline{88.49} & \underline{42.85} & \underline{58.36} & 91.36 & \underline{83.46} & \textbf{47.95} \\


MM++ (Ours) & \underline{91.34} & \underline{43.20} & \underline{96.43} & \underline{20.98} & \textbf{84.35} & \underline{53.80} & \textbf{81.70} & \textbf{52.43} & \textbf{88.84} & \textbf{41.83} & \textbf{60.78} & \textbf{88.94} & \textbf{83.91} & \underline{50.20} \\

\bottomrule
\end{tabular}%
}
\end{table*}

\begin{figure}[htb]
    \vspace{-0.25cm}
    \centering
    \includegraphics[width=\linewidth]{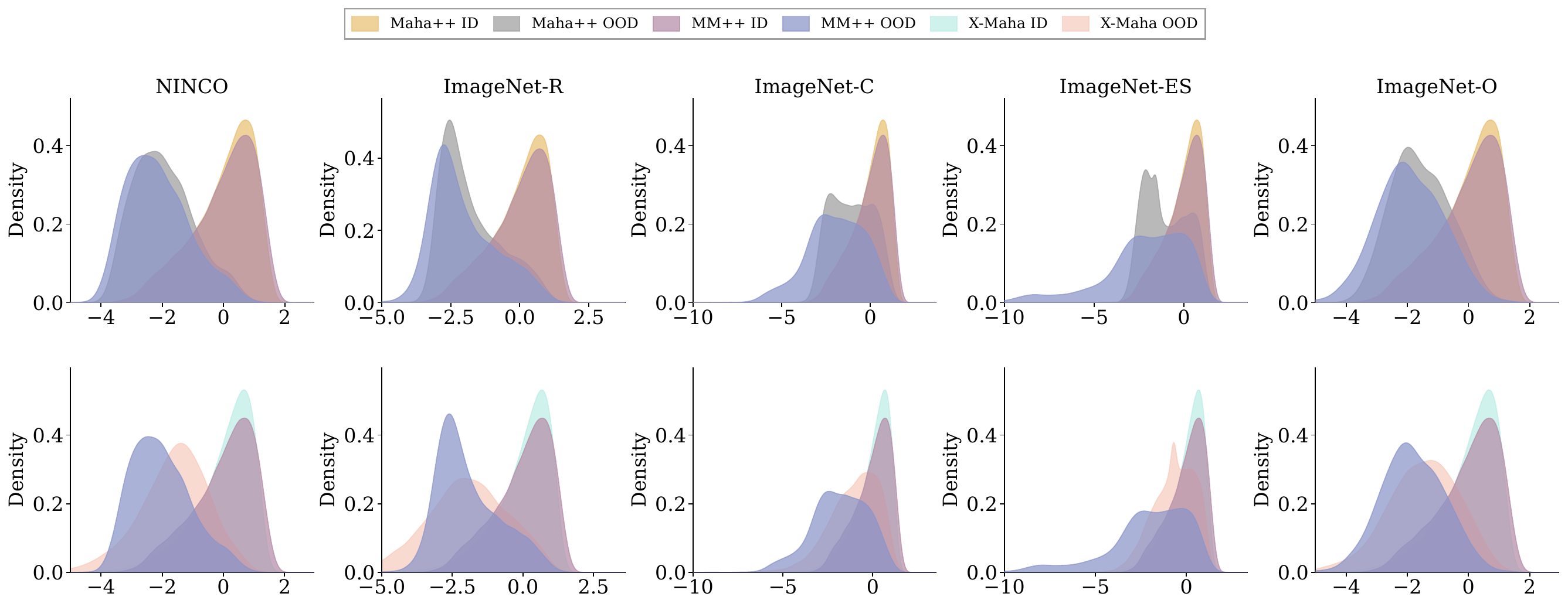}
    \caption{Score distributions for ViT-B/16 across five OOD benchmarks with ImageNet-LT as ID.}
    \label{fig:ood_score_histograms_imagenetlt_vitb16}
    \vspace{-0.25cm}
\end{figure}

\begin{figure}[htbp]
    \vspace{-0.25cm}
    \centering
    \includegraphics[width=\textwidth]{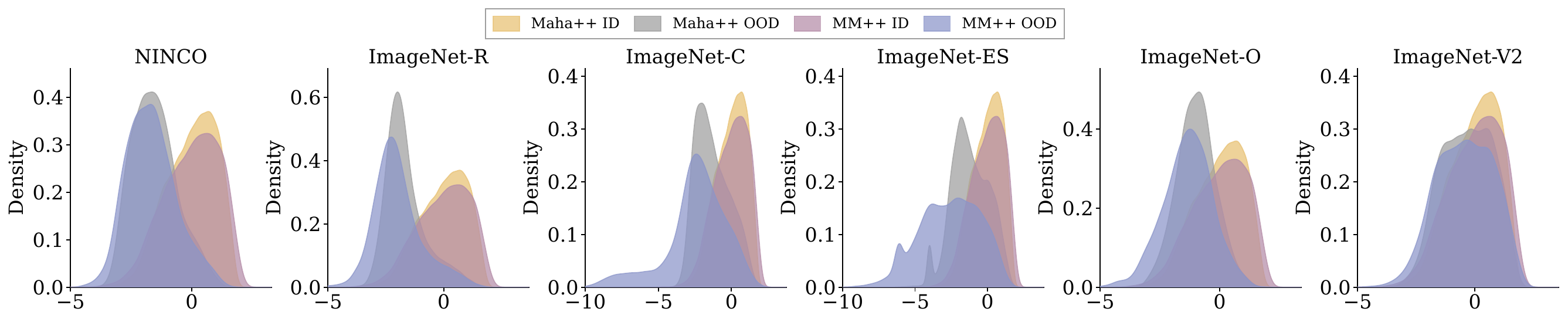}
    \caption{Score distributions for Swin-T across six OOD benchmarks with {ImageNet-LT} as ID.}
    \label{fig:imgnetlt_ood_score_distribution_swint}
    \vspace{-0.25cm}
\end{figure}

\textbf{Analysis of Score Distributions (ViT-B/16 and Swin-T).} Figures~\ref{fig:ood_score_histograms_imagenetlt_vitb16} and~\ref{fig:imgnetlt_ood_score_distribution_swint} illustrate score distributions under the ImageNet-LT ID regime. Across both the ViT-B/16 and hierarchical Swin-T backbones, MM++ effectively handles distinct shift types. Under continuous covariate shifts (ImageNet-C, -ES), MM++ exhibits a pronounced long tail (extending near $-10$), demonstrating fine-grained sensitivity to degradation severity. Conversely, for pure semantic (NINCO, ImageNet-O) and severe stylistic shifts (ImageNet-R), MM++ forms a concentrated, isolated mass, effectively separating these conceptual anomalies from the ID distribution.

Comparatively, the single-layer Mahalanobis++ baseline produces tightly clustered scores with substantial ID overlap on covariate shifts. X-Mahalanobis incorporates multi-layer data but falls short of MM++’s sensitivity on ViT-B/16, likely due to its additive fusion which aggregates independent layer scores rather than constructing a unified feature space. Furthermore, X-Mahalanobis is excluded from the Swin-T evaluation; its assumption of homogeneous feature dimensionality makes it incompatible with Swin-T's stage-wise spatial compression and channel expansion. 

Ultimately, this cross-architectural consistency highlights MM++’s core advantage: by leveraging top-$K$ information gating and constructing a unified representation space, the framework seamlessly captures multi-layer correlations across distinct attention paradigms without requiring architecture-specific dimensional adaptations.

\begin{table}[t]
\centering
\caption{Impact of layer selection and fusion strategies in MM++ on OOD detection with ViT-B/16, evaluated on ImageNet-C (ImageNet-LT as ID).}
\label{tab:ablation}
\vspace{0.15cm}
\resizebox{0.8\textwidth}{!}{%
\begin{tabular}{lcc}
\toprule
Aggregation Strategy & AUROC $\uparrow$ & FPR95 $\downarrow$ \\
\midrule
\multicolumn{3}{l}{\textit{Single-layer baselines}} \\
Penultimate layer (Mahalanobis) & 70.94 & 85.00 \\
Penultimate layer + $\ell_2$-norm (Mahalanobis++) & 74.86 & 73.05\\
\midrule
\multicolumn{3}{l}{\textit{Multi-layer fusion (prior / supervised)}} \\
Variance-based weighting + fine-tuning (X-Mahalanobis$^\dagger$) & \underline{83.96} & \textbf{53.57} \\
\midrule
\multicolumn{3}{l}{\textit{MM++ design variants}} \\
(1) Top-$K$ layers + no penultimate anchoring + pseudo-inverse & 73.47 & 83.53 \\
(2) Top-$K$--1 layers + penultimate anchoring + pseudo-inverse  
& 76.09 & 71.30 \\ \midrule
MM++ (Top-$K$--1 + penultimate + Ledoit-Wolf shrinkage) & \textbf{84.35} & \underline{53.80} \\
\bottomrule
\end{tabular}
}
\vspace{-0.55cm} 
\end{table}

\subsection{Ablation Studies}
\label{sec:ablations}

Table~\ref{tab:ablation} details the effect of progressive layer aggregation on ViT-B/16, evaluated on ImageNet-C (ImageNet-LT ID).

\textbf{(1) Single-layer limitations.} Relying solely on the penultimate layer (Mahalanobis) yields moderate performance, as a single representation fails to capture structural deviations critical for near-OOD detection. Incorporating $\ell_2$ normalization (Mahalanobis++) improves performance, highlighting the necessity of scale invariance in high-dimensional spaces.

\textbf{(2) Multi-layer fusion improves separation.} Extending beyond single representations significantly enhances detection. In particular, X-Mahalanobis achieves competitive results via variance-based weighting and fine-tuning. However, their heuristic aggregation introduces sensitivity to less informative layers and requires supervised calibration, limiting strictly post-hoc practicality.

\textbf{(3) Anchoring the terminal representation.} Selecting layers solely via entropy density drops without enforcing penultimate inclusion causes substantial degradation (particularly in FPR95) compared to the anchored MM++ variant. This shows the penultimate layer provides a necessary structural anchor; retaining it while selecting the top $K-1$ intermediate layers ensures stable discrimination.

\begin{wrapfigure}{r}{0.4\linewidth}
    \vspace{-0.25cm}
    \centering
    \includegraphics[width=\linewidth]{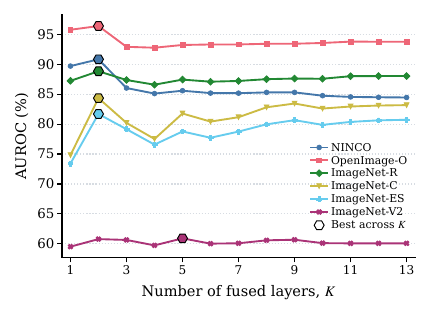}
    \caption{Sensitivity of MM++ to $K$
    }
    \label{fig:topk_ablation}
    \vspace{-0.5cm}
\end{wrapfigure}

\textbf{(4) Robust precision estimation.} While traditional methods utilize pseudo-inverse estimation for the tied precision matrix, MM++ adopts Ledoit–Wolf shrinkage for well-conditioned covariance. This numerically stable foundation yields substantial empirical gains—an 8.26\% AUROC and 17.5\% FPR95 improvement over the pseudo-inverse baseline. Crucially, this effectiveness is intrinsically linked to our top-$K$ gating, which isolates semantically relevant signals for the shrinkage estimator (see Table~\ref{tab:maha_cov_ablation} in Appendix~\ref{app:more-results}).


\textbf{(5) Sensitivity to capacity parameter $K$.} MM++ exhibits measured sensitivity to $K$, as shown in Figure~\ref{fig:topk_ablation} (ViT-B/16, ImageNet-LT as ID). AUROC jumps significantly from $K=1$ to $K=2$, affirming that intermediate information crucially complements the terminal representation. However, performance fluctuates, dropping at $K=3$ before rising for larger $K$. Structurally, this non-monotonicity stems from information redundancy: subsequently ranked layers likely capture highly correlated features. At $K=3$, marginal information gains are temporarily outweighed by the statistical noise of expanding the joint dimensionality. While $K=5$ marginally outperforms $K=2$ on ImageNet-V2 by eventually accumulating orthogonal signals, the minor improvement does not justify the increased computational footprint. Ultimately, small $K$ values provides the best trade-off between performance and efficiency.



%% file: 5.conc.tex
\vspace{-0.2cm}
\section{Conclusion}
\label{sec:conclusion}
\vspace{-0.2cm}

In this work, we introduce MM++, a fully unsupervised framework for post-hoc Out-of-Distribution (OOD) detection. Our approach moves beyond reliance on terminal representations by demonstrating that the hierarchical trajectory of feature evolution contains valuable discriminative information.
The effectiveness of MM++ stems from modeling hierarchical patterns without requiring heuristic weighting or auxiliary OOD datasets. We propose a principled method that identifies layers where semantic compression is most pronounced via entropy density drops. By fusing the anchored penultimate layer with these selected intermediate representations into a joint Mahalanobis++ space, the framework captures cross-layer correlations latent in single-layer approaches. This effectively addresses the dichotomy between the stability of terminal layers and the depth of multilayer architectures.
Ultimately, the consistent performance of MM++ across representative Transformer and Convolutional models—specifically ViT, EVA02, Swin, and ConvNeXt—suggests that robust OOD detection can be achieved through generalizable principles rather than architecture-specific tuning. By grounding detection in information-theoretic and geometric foundations, MM++ offers a rigorous path toward more reliable deep learning systems.

%% file: 7.Appendix.tex
\appendix

\input{App-justify3}

\section{Additional Results}
\label{app:more-results}

\subsection{AUROC Visualization}

\begin{figure}[htbp]
  \centering
  \begin{subfigure}[b]{0.48\textwidth}
    \centering
    \includegraphics[width=\linewidth]{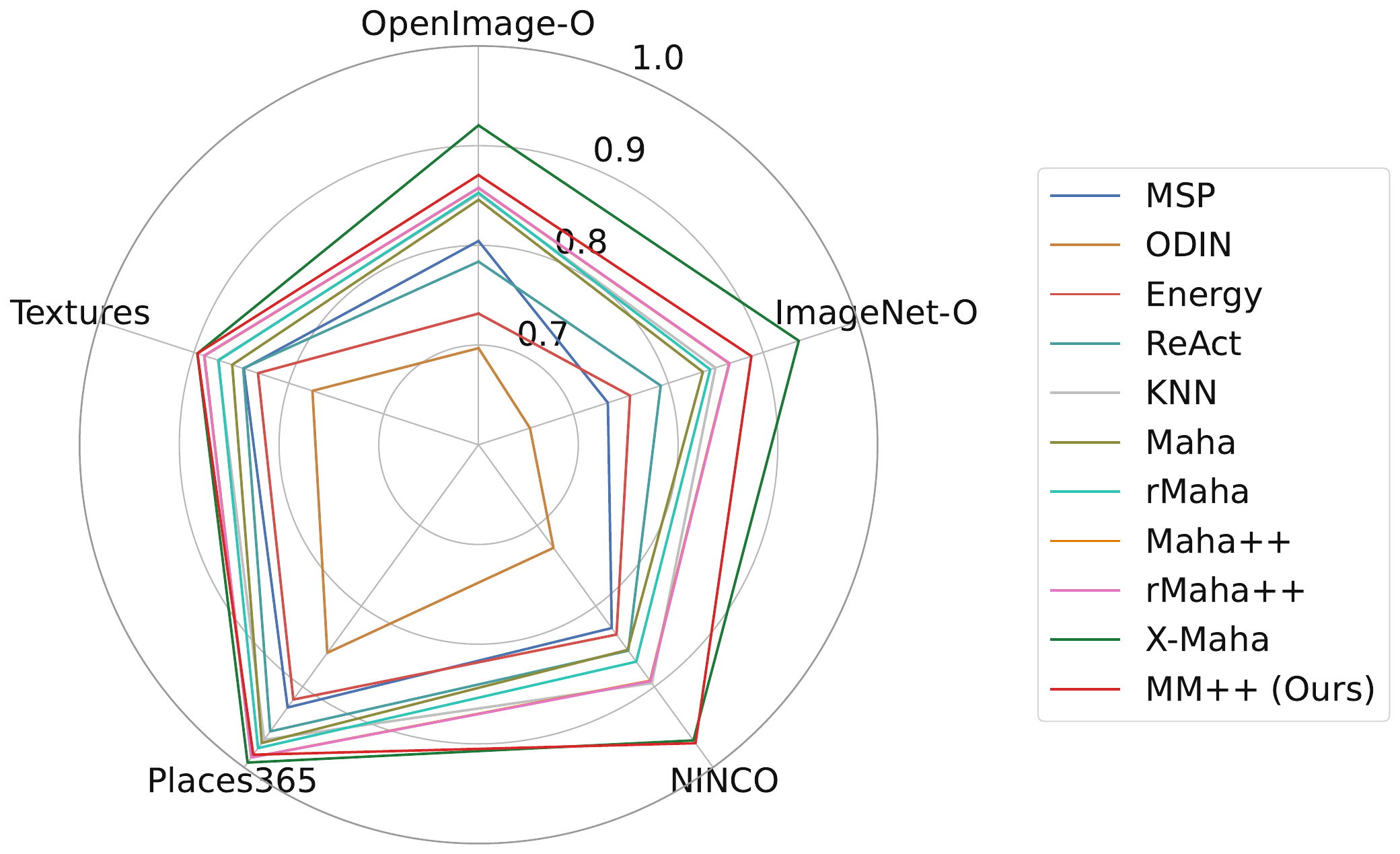}
    \caption{ID dataset: Imagenet-1K, Model: ViT-B/16}
    \label{fig:sub1}
  \end{subfigure}
  \hfill 
  \begin{subfigure}[b]{0.48\textwidth}
    \centering
    \includegraphics[width=\linewidth]{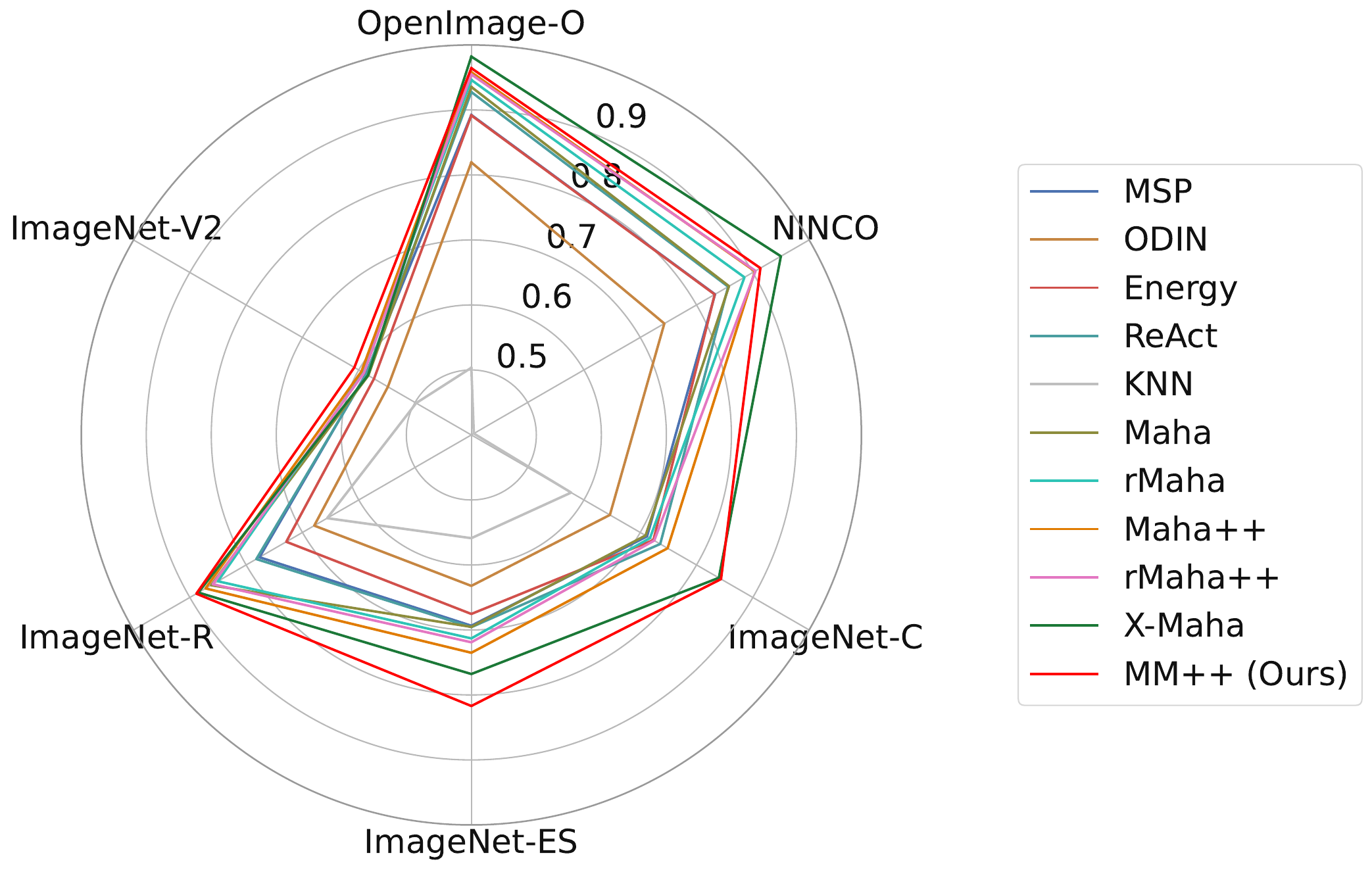}
    \caption{ID dataset: Imagenet-LT, Model: ViT-B/16}
    \label{fig:sub2}
  \end{subfigure}
  \caption{AUROC comparison across datasets, visualizing the results in Tables~\ref{tab:main_results} and~\ref{tab:imagenetlt_results}}
  \label{fig:radar_imagenet1k_lt}
\end{figure}

To provide a holistic view of OOD detection performance across diverse distribution shifts, Figure~\ref{fig:radar_imagenet1k_lt}(a) and~(b) visualize the AUROC scores of MM++ against a comprehensive suite of baseline methods using the ViT-B/16 architecture.

\textbf{Performance on Balanced ID (ImageNet-1K).} Figure~\ref{fig:radar_imagenet1k_lt}(a) evaluates robustness under the standard ImageNet-1K regime. As illustrated, X-Mahalanobis consistently forms the outermost perimeter of the radar plot, demonstrating its effectiveness for a balanced ID. It outperforms the other approaches particularly on far-OOD benchmarks likely due to its fine-tuning. MM++ is competitive on far-OOD datasets, while achieving the best AUROC on Texture and SUN. Therefore, this visually confirms that MM++ is a fully post-hoc, unsupervised alternative.

\textbf{Robustness under Class Imbalance (ImageNet-LT).} The advantages of MM++ are even more pronounced under the challenging long-tailed regime as illustrated in Figure~\ref{fig:radar_imagenet1k_lt}(b). While traditional distance-based methods 
suffer from unreliable statistic estimation on data-poor tail classes, MM++ remains much more stable. It establishes a dominant perimeter across all evaluated near-OOD benchmarks. This sustained resilience highlights the efficacy of combining top-$K$ feature gating with shrinkage-based covariance calibration in the fused space, effectively mitigating the geometric vulnerabilities associated with underrepresented tail classes.

Collectively, these visualizations demonstrate that unlike single-layer baselines or additive multi-layer approaches (e.g., X-Mahalanobis), MM++ avoids dataset-specific failure modes, establishing a universally robust framework for varied deployment conditions.

\subsection{Analysis of MM++ Sensitivity to $K$}

\begin{figure}[htb]
    \centering
    \begin{subfigure}[t]{0.48\linewidth}
      \centering
      \includegraphics[width=\linewidth]{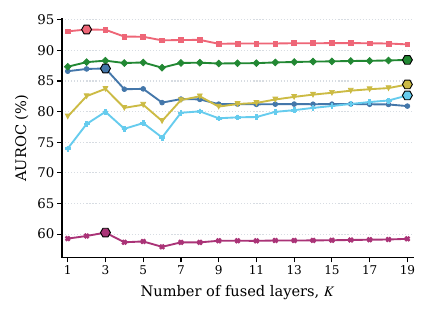}
      \caption{ConvNeXt-T}
      \label{fig:k-sens-convnext}
    \end{subfigure}
    \begin{subfigure}[t]{0.48\linewidth}
      \centering
      \includegraphics[width=\linewidth]{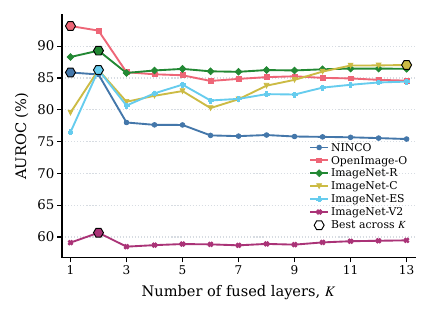}
      \caption{Swin-T}
      \label{fig:k-sens-swin}
    \end{subfigure}

    \caption{AUROC of MM++ as a function of the number of fused layers $K \in {1, \dots, k_{max}}$ for each model, where $K=1$ corresponds to using only the penultimate layer.}
    \label{fig:k-sensitivity-imagenetlt}
  \end{figure}

Figure~\ref{fig:k-sensitivity-imagenetlt} illustrates MM++'s sensitivity to $K$ on ImageNet-LT as the ID dataset. As discussed in Section~\ref{sec:eval}, a peak AUROC is typically achieved for a small $K$ value. Subsequently, AUROC drops until it rises again for large $K$ values. Nonetheless, the empirical AUROC of higher $K$ values are lower than the peak across diverse OOD datasets. Therefore, this demonstrates that a small $K$ value (e.g., $K=2$) provides best trade-offs between performance and complexity, conforming to the ablation results discussed in Section~\ref{sec:eval}.

\subsection{EVA02-S14 Results}
Table~\ref{tab:imagenetlt_eva02_s14_results} presents results on ImageNet-LT using the EVA02-S14 backbone (\texttt{eva02\_small\_patch14\_336.mim\_in22k\_ft\_in1k}). 
As expected, confidence-based methods degrade under long-tailed distributions, while Mahalanobis-based approaches provide stronger baselines.

A clear distinction emerges between near- and far-OOD settings. Single-layer methods and X-Mahalanobis perform well on far-OOD datasets, where large semantic shifts are present. In contrast, MM++ consistently achieves the best performance on near-OOD benchmarks (ImageNet-C, ImageNet-ES, ImageNet-R, and ImageNet-V2).

This behavior highlights the key advantage of MM++: near-OOD samples induce subtle, distributed deviations across multiple layers, which cannot be captured by single-layer representations or independent score aggregation. By modeling cross-layer dependencies, MM++ provides a more sensitive and robust detection mechanism for realistic distribution shifts.

\subsection{ConvNeXt-T Results}
\begin{figure}[htb]
    \centering
    \includegraphics[width=\textwidth]{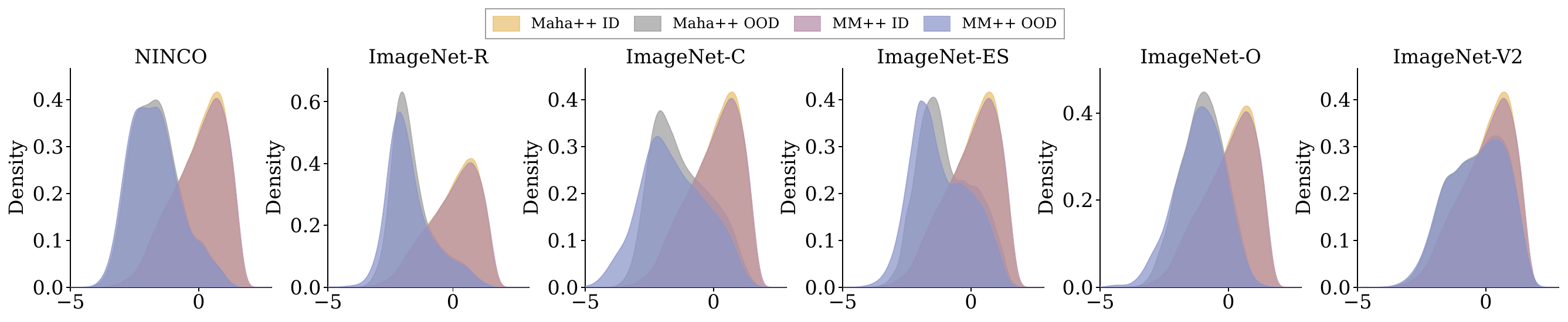}
    \caption{Score distributions for ConvNeXt-T across six OOD benchmarks with {ImageNet-LT} as ID.}
    \label{fig:imgnetlt_score_distribution_convnext_t}
\end{figure}

Table~\ref{tab:convnext_t_results} and Figure \ref{fig:imgnetlt_score_distribution_convnext_t} shows the results, and  visualizes the score distribution for the ConvNeXt-T backbone (\texttt{convnext\_tiny.fb\_in1k}). 
The similar trend persists across architectures. MM++ consistently outperforms all baselines on near-OOD datasets and achieves the best overall performance. 

Unlike ViT and EVA02-S14, X-Mahalanobis is not applicable in this setting due to its reliance on homogeneous transformer features and adaptation-based design. This highlights an important advantage of MM++: it operates purely in a post-hoc manner and naturally extends to heterogeneous architectures without requiring model modification.

Overall, the results demonstrate that multi-layer fusion enables MM++ to capture both low-level covariate shifts and high-level semantic inconsistencies, leading to improved robustness under long-tailed conditions.

\begin{table*}[htb]
\centering
\caption{OOD detection performance using EVA02-S14 with ImageNet-LT as ID.}
\label{tab:imagenetlt_eva02_s14_results}
\small
\setlength{\tabcolsep}{2pt}
\renewcommand{\arraystretch}{1.9}
\resizebox{\textwidth}{!}{%
\begin{tabular}{>{\LARGE}l >{\LARGE}c>{\LARGE}c >{\LARGE}c>{\LARGE}c >{\LARGE}c>{\LARGE}c >{\LARGE}c>{\LARGE}c >{\LARGE}c>{\LARGE}c >{\LARGE}c>{\LARGE}c >{\LARGE}c>{\LARGE}c}
\toprule
\multirow{2}{*}{\textbf{Method}} & \multicolumn{2}{c}{\Large \textbf{NINCO}} &
\multicolumn{2}{c}{\Large \textbf{OpenImage-O}} &
\multicolumn{2}{c}{\Large \textbf{ImageNet-C}} &
\multicolumn{2}{c}{\Large \textbf{ImageNet-ES}} &
\multicolumn{2}{c}{\Large \textbf{ImageNet-R}} &
\multicolumn{2}{c}{\Large \textbf{ImageNet-V2}} & \multicolumn{2}{c}{\Large \textbf{Average}} \\
& {\large AUROC $\uparrow$} & {\large FPR95 $\downarrow$}
& {\large AUROC $\uparrow$} & {\large FPR95 $\downarrow$}
& {\large AUROC $\uparrow$} & {\large FPR95 $\downarrow$}
& {\large AUROC $\uparrow$} & {\large FPR95 $\downarrow$}
& {\large AUROC $\uparrow$} & {\large FPR95 $\downarrow$}
& {\large AUROC $\uparrow$} & {\large FPR95 $\downarrow$}
& {\large AUROC $\uparrow$} & {\large FPR95 $\downarrow$} \\  
\midrule
MSP & 80.99 & 62.28 & 89.49 & 41.19 & 70.32 & 72.77 & 67.84 & 70.91 & 77.35 & 61.66 & 57.82 & 89.82 & 73.97 & 66.44 \\
ODIN & 71.31 & 72.80 & 82.24 & 52.54 & 65.38 & 78.96 & 62.65 & 80.21 & 70.52 & 70.99 & 55.47 & 91.79 & 67.93 & 74.55 \\
Energy & 75.62 & 62.35 & 86.28 & 39.72 & 66.54 & 74.06 & 65.32 & 73.25 & 74.11 & 58.17 & 55.87 & 90.43 & 70.62 & 66.33 \\
ReAct & 76.17 & 62.35 & 86.75 & 39.01 & 66.34 & 74.44 & 65.29 & 73.45 & 74.42 & 57.91 & 55.94 & 90.37 & 70.82 & 66.25 \\
KNN & 72.75 & 80.39 & 87.05 & 53.32 & 61.06 & 90.85 & 58.05 & 92.50 & 75.06 & 64.85 & 53.92 & 92.32 & 67.98 & 79.04 \\
\midrule
Maha & 87.95 & 51.89 & \underline{95.92} & \underline{21.36} & 74.25 & 68.46 & 70.52 & 69.66 & 85.13 & 48.55 & 58.16 & 89.27 & 78.65 & 58.20 \\
rMaha & 88.28 & \underline{50.94} & 95.54 & 22.96 & 73.90 & 69.21 & 70.35 & 68.08 & 84.15 & 51.24 & \underline{58.44} & 89.27 & 78.44 & 58.62 \\
Maha++ & \underline{88.42} & \textbf{50.29} & \textbf{96.08} & \textbf{20.35} & 75.20 & 65.82 & 71.42 & 65.45 & 85.39 & 47.66 & 58.35 & \underline{89.24} & 79.15 & \textbf{56.47} \\
rMaha++ & 88.42 & \textbf{50.29} & \textbf{96.08} & \textbf{20.35} & 75.20 & \underline{65.81} & 71.42 & \underline{65.45} & 85.39 & 47.66 & 58.36 & \underline{89.24} & 79.15 & \textbf{56.47} \\
\midrule
$\text{X-Maha}^{\dagger}$ & \textbf{88.71} & 51.75 & 95.50 & 24.17 & \underline{76.70} & 65.17 & \underline{74.30} & 64.14 & \underline{87.15} & \underline{46.55} & 58.12 & 90.09 & \underline{80.08} & \underline{56.98} \\


MM++ (Ours) & 82.06 & 72.87 & 87.49 & 66.96 & \textbf{89.25} & \textbf{42.51} & \textbf{88.24} & \textbf{47.31} & \textbf{90.31} & \textbf{42.91} & \textbf{65.68} & \textbf{85.62} & \textbf{83.84} & 59.70 \\
\bottomrule
\end{tabular}%
}
\vspace{+0.3cm}

\caption{OOD detection performance using ConvNeXt-T with ImageNet-LT as ID.}
\label{tab:convnext_t_results}
\small
\setlength{\tabcolsep}{2pt}
\renewcommand{\arraystretch}{1.9}
\resizebox{\textwidth}{!}{%
\begin{tabular}{>{\LARGE}l >{\LARGE}c>{\LARGE}c >{\LARGE}c>{\LARGE}c >{\LARGE}c>{\LARGE}c >{\LARGE}c>{\LARGE}c >{\LARGE}c>{\LARGE}c >{\LARGE}c>{\LARGE}c >{\LARGE}c>{\LARGE}c}
\toprule
\multirow{2}{*}{\textbf{Method}} &
\multicolumn{2}{c}{\Large \textbf{NINCO}} &
\multicolumn{2}{c}{\Large \textbf{OpenImage-O}} &
\multicolumn{2}{c}{\Large \textbf{ImageNet-C}} &
\multicolumn{2}{c}{\Large \textbf{ImageNet-R}} &
\multicolumn{2}{c}{\Large \textbf{ImageNet-ES}} &
\multicolumn{2}{c}{\Large \textbf{ImageNet-V2}} &
\multicolumn{2}{c}{\Large \textbf{Average}} \\
& {\large AUROC $\uparrow$} & {\large FPR95 $\downarrow$}
& {\large AUROC $\uparrow$} & {\large FPR95 $\downarrow$}
& {\large AUROC $\uparrow$} & {\large FPR95 $\downarrow$}
& {\large AUROC $\uparrow$} & {\large FPR95 $\downarrow$}
& {\large AUROC $\uparrow$} & {\large FPR95 $\downarrow$}
& {\large AUROC $\uparrow$} & {\large FPR95 $\downarrow$}
& {\large AUROC $\uparrow$} & {\large FPR95 $\downarrow$}\\
\midrule
MSP      & 80.68 & 65.45 & 82.83 & 61.63 & 78.65 & \underline{60.80} & 77.58 & 67.02 & \underline{74.36} & \textbf{65.06} & \textbf{61.53} & \textbf{88.49} & 75.94 & 68.08 \\
ODIN     & 61.74 & 88.38 & 58.64 & 89.85 & 69.32 & 80.01 & 55.07 & 92.33 & 64.60 & 78.82 & 59.71 & 90.68 & 61.51 & 86.68 \\
Energy   & 63.71 & 87.11 & 57.11 & 92.83 & 70.11 & 80.46 & 57.71 & 91.89 & 64.98 & 76.09 & 58.86 & 89.73 & 62.08 & 86.35 \\
ReAct    & 68.02 & 85.23 & 64.45 & 90.72 & 73.33 & 77.46 & 64.86 & 89.88 & 67.80 & 75.12 & \underline{60.06} & \underline{89.37} & 66.42 & 84.63 \\
KNN      & 34.64 & 99.07 & 44.80 & 97.76 & 59.54 & 84.86 & 65.56 & 91.85 & 52.80 & 94.42 & 46.21 & 96.34 & 50.59 & 94.05 \\
\midrule
Maha     & 82.11 & 74.49 & 91.57 & 48.96 & 75.59 & 74.85 & 87.03 & \underline{51.92} & 70.13 & 82.78 & 57.61 & 91.22 & 77.34 & 70.70 \\
rMaha    & 85.43 & 66.01 & 92.61 & 42.71 & 77.74 & 69.75 & 85.55 & 52.62 & 73.51 & 79.57 & 59.40 & 90.43 & 79.04 & 66.85 \\
Maha++   & 86.62 & 56.49 & 93.12 & 39.51 & \underline{79.27} & 64.26 & \underline{87.32} & 52.25 & 74.33 & 78.41 & 59.29 & 90.32 & 79.99 & 63.54 \\
rMaha++  & \underline{86.68} & \underline{56.45} & \underline{93.14} & \underline{39.47} & 79.24 & 64.32 & 87.29 & 52.37 & 74.35 & 78.48 & 59.32 & 90.32 & 80.00 & 63.57 \\
\midrule
MM++ (Ours) & \textbf{86.97} & \textbf{55.61} & \textbf{93.40} & \textbf{37.69} & \textbf{82.52} & \textbf{56.39} & \textbf{88.08} & \textbf{49.12} & \textbf{77.98} & \underline{69.36} & 59.70 & 90.33 & \textbf{81.44} & \textbf{59.75} \\
\bottomrule
\end{tabular}%
}
\vspace{+0.3cm}
\caption{OOD detection performance using Swin-T with ImageNet-LT as ID.}
\label{tab:swin_t_results}
\small
\setlength{\tabcolsep}{2pt}
\renewcommand{\arraystretch}{1.9}
\resizebox{\textwidth}{!}{%
\begin{tabular}{>{\LARGE}l >{\LARGE}c>{\LARGE}c >{\LARGE}c>{\LARGE}c >{\LARGE}c>{\LARGE}c >{\LARGE}c>{\LARGE}c >{\LARGE}c>{\LARGE}c >{\LARGE}c>{\LARGE}c >{\LARGE}c>{\LARGE}c}
\toprule
\multirow{2}{*}{\textbf{Method}} &
\multicolumn{2}{c}{\Large \textbf{NINCO}} &
\multicolumn{2}{c}{\Large \textbf{OpenImage-O}} &
\multicolumn{2}{c}{\Large \textbf{ImageNet-C}} &
\multicolumn{2}{c}{\Large \textbf{ImageNet-R}} &
\multicolumn{2}{c}{\Large \textbf{ImageNet-ES}} &
\multicolumn{2}{c}{\Large \textbf{ImageNet-V2}} &
\multicolumn{2}{c}{\Large \textbf{Average}} \\
& {\large AUROC $\uparrow$} & {\large FPR95 $\downarrow$}
& {\large AUROC $\uparrow$} & {\large FPR95 $\downarrow$}
& {\large AUROC $\uparrow$} & {\large FPR95 $\downarrow$}
& {\large AUROC $\uparrow$} & {\large FPR95 $\downarrow$}
& {\large AUROC $\uparrow$} & {\large FPR95 $\downarrow$}
& {\large AUROC $\uparrow$} & {\large FPR95 $\downarrow$}
& {\large AUROC $\uparrow$} & {\large FPR95 $\downarrow$}\\
\midrule
MSP      & 80.30 & 70.52 & 85.68 & 60.21 & 76.19 & 68.36 & 80.14 & 63.92 & 76.17 & 64.88 & 59.86 & 90.11 & 76.39 & 69.67 \\
ODIN     & 67.68 & 85.41 & 72.12 & 81.28 & 62.81 & 85.08 & 72.10 & 78.45 & 63.80 & 80.89 & 56.70 & 92.02 & 65.87 & 83.86 \\
Energy   & 77.10 & 72.42 & 79.08 & 67.98 & 75.27 & 65.33 & 76.95 & 66.59 & 76.74 & 59.48 & 59.77 & 90.18 & 74.15 & 70.33 \\
ReAct    & 78.62 & 72.22 & 82.93 & 64.08 & 76.46 & 64.10 & 80.74 & 60.12 & 77.53 & 59.39 & \underline{60.40} & \underline{89.88} & 76.11 & 68.30 \\
KNN      & 43.16 & 95.30 & 54.11 & 89.41 & 66.16 & 74.15 & 77.77 & 65.05 & 61.98 & 82.33 & 50.56 & 93.93 & 58.96 & 83.36 \\
\midrule
Maha     & 80.46 & 79.76 & 90.47 & 54.30 & 74.54 & 77.28 & 87.10 & 52.10 & 67.36 & 88.58 & 57.86 & 91.30 & 76.30 & 73.89 \\
rMaha    & 83.59 & 70.50 & 91.87 & 43.74 & 76.37 & 70.69 & 85.45 & 52.49 & 73.12 & 83.09 & 58.56 & 90.49 & 78.16 & 68.50 \\
Maha++   & \underline{85.87} & \underline{57.79} & \underline{93.14} & \textbf{36.86} & 79.60 & 58.98 & \underline{88.32} & \underline{45.72} & 76.63 & \underline{63.90} & 59.13 & 89.92 & 80.45 & 58.86 \\
rMaha++  & \textbf{86.32} & \textbf{56.21} & \textbf{93.15} & \underline{37.03} & \underline{80.14} & \underline{57.97} & 88.29 & 46.65 & \underline{77.38} & 62.67 & 59.33 & 89.95 & \underline{80.77} & \underline{58.41} \\
\midrule
MM++ (Ours) & 85.55 & 60.07 & 92.44 & 43.12 & \textbf{86.26} & \textbf{43.56} & \textbf{89.29} & \textbf{42.91} & \textbf{86.25} & \textbf{41.09} & \textbf{60.67} & \textbf{89.61} & \textbf{83.41} & \textbf{53.39} \\
\bottomrule
\end{tabular}%
}
\vspace{-0.5cm}
\end{table*}

\subsection{Swin-T Results}
Table~\ref{tab:swin_t_results} reports results using the Swin-T backbone (\texttt{swin\_tiny\_patch4\_window7\_224.ms\_in1k}). 
The observations remain consistent with the previous settings.

MM++ achieves the strongest performance on all near-OOD benchmarks, while remaining competitive on far-OOD datasets. In contrast, single-layer Mahalanobis variants show stronger performance on large semantic shifts but struggle to capture subtle distribution changes.

Similar to ConvNeXt-T, X-Mahalanobis is not applicable to Swin-T, further emphasizing the architectural flexibility of MM++. These results confirm that modeling cross-layer interactions is particularly beneficial for detecting structured covariate shifts in modern hierarchical architectures.

\begin{table*}[htb]
\centering
\caption{AUROC on Near-OOD datasets averaged over ImageNet-V2, ImageNet-C, ImageNet-R, and ImageNet-ES with ImageNet-LT (ID)} 
\label{tab:near_ood_shift_auroc}
\renewcommand{\arraystretch}{1.3}
\resizebox{\textwidth}{!}{%
\begin{tabular}{>{\large}l >{\large}c >{\large}c >{\large}c >{\large}c >{\large}c >{\large}c >{\large}c >{\large}c >{\large}c >{\large}c >{\large}c}
\toprule
\large Model & \makecell{\normalsize Val\\Acc} & \normalsize MSP & \normalsize ODIN & \normalsize Energy & \normalsize ReAct & \normalsize KNN & \normalsize Maha & \normalsize rMaha & \normalsize Maha++ & \normalsize rMaha++ & \makecell{\normalsize MM++\\(ours)} \\

\midrule
ViT-B16-In21k-augreg & 84.47 & 69.50 & 72.30 & 73.39 & 73.17 & 61.80 & 73.60 & 72.84 & \underline{73.73} & 73.72 & \textbf{80.15} \\
ViT-L16-In21k-augreg & 85.81 & 69.63 & 74.39 & 74.29 & 74.20 & 61.36 & \underline{74.64} & 72.46 & 74.48 & 74.47 & \textbf{78.98} \\
ViT-T16-In21k-augreg & 75.48 & 73.01 & 72.32 & 77.46 & 76.78 & 71.17 & 69.52 & 72.96 & \underline{77.55} & 77.53 & \textbf{81.26} \\
ViT-S16-In21k-augreg & 81.37 & 70.84 & 72.72 & 75.33 & 75.05 & 64.92 & 72.05 & 72.78 & \underline{75.67} & 75.64 & \textbf{82.62} \\
ViT-B16-augreg & 79.13 & 71.27 & 70.01 & 74.27 & 74.10 & 54.89 & 73.77 & 73.15 & \underline{74.31} & 74.29 & \textbf{74.48} \\
ViT-S16-augreg & 78.79 & 71.74 & 70.11 & 74.37 & \underline{74.45} & 47.85 & 74.19 & 74.02 & 74.32 & 74.30 & \textbf{80.53} \\
ViT-so400M-SigLip & 89.33 & 64.44 & 61.14 & 62.08 & 64.38 & 55.08 & 70.53 & 68.90 & \textbf{71.94} & \underline{71.93} & 66.22 \\
ViT-B16-In21k-orig & 81.80 & 71.17 & 70.33 & 73.89 & 73.83 & 56.45 & 73.85 & 73.76 & \underline{74.50} & \textbf{74.58} & 74.24 \\
ViT-L16-In21k-orig & 81.50 & 71.16 & 69.54 & 73.02 & 73.08 & 53.77 & 73.18 & 73.81 & \textbf{74.44} & \underline{74.43} & 73.58 \\
ViT-B16-In21k-mil & 84.26 & 71.87 & 70.73 & 73.66 & \underline{73.75} & 53.87 & 70.01 & 72.33 & 73.68 & \textbf{73.93} & 70.34 \\
ViT-B16-In21k-augreg2 & 85.06 & 69.23 & 64.04 & 67.51 & 69.99 & 57.26 & 71.25 & 71.81 & 73.78 & \underline{73.82} & \textbf{78.92} \\
EVA02-L14-M38m-In21k & 89.56 & 62.30 & 63.62 & 60.81 & 61.05 & 47.03 & 67.88 & 66.36 & 68.83 & \underline{68.83} & \textbf{81.68} \\
EVA02-B14-In21k & 88.38 & 65.44 & 63.17 & 62.97 & 63.78 & 45.22 & 69.76 & 68.89 & 71.04 & \underline{71.04} & \textbf{82.11} \\
EVA02-S14 & 85.64 & 68.33 & 63.51 & 65.46 & 65.50 & 62.02 & 72.02 & 71.71 & 72.59 & \underline{72.59} & \textbf{83.37} \\
EVA02-T14 & 80.58 & 70.11 & 64.20 & 68.90 & 68.83 & 63.79 & 73.21 & 72.91 & 74.02 & \underline{74.03} & \textbf{78.02} \\
DeiT3-B16 & 83.72 & 69.83 & 65.15 & 67.92 & 66.67 & 67.04 & 71.51 & 71.40 & \underline{72.11} & \textbf{72.33} & 71.90 \\
DeiT3-L16-In21k & 87.58 & 65.99 & 65.51 & 65.04 & 67.55 & 57.84 & 70.38 & 68.06 & \underline{70.94} & \textbf{71.02} & 70.78 \\
DeiT3-L16 & 85.80 & 66.25 & 64.41 & 63.54 & 56.29 & 69.69 & 72.35 & 70.72 & \textbf{72.88} & 71.67 & \underline{72.60} \\
DeiT3-B16-In1k & 84.95 & 70.33 & 61.91 & 64.35 & 60.43 & 68.66 & 72.58 & 72.08 & 73.11 & \textbf{73.39} & \underline{73.14} \\
DeiT3-B16-384-In1k & 86.67 & 66.83 & 62.55 & 62.81 & 67.06 & 49.80 & 72.23 & 70.89 & \underline{72.88} & \textbf{72.89} & 72.75 \\
DeiT3-B16-224-In1k & 85.72 & 66.95 & 63.70 & 63.83 & 67.57 & 51.36 & 71.64 & 70.53 & \underline{72.39} & \textbf{72.40} & 72.19 \\
DeiT3-S16-In21k & 84.74 & 66.46 & 59.18 & 64.66 & 66.59 & 42.78 & 71.78 & 71.33 & \underline{72.61} & \textbf{72.61} & 72.11 \\
DeiT3-S16 & 83.44 & 71.60 & 64.41 & 68.35 & 67.42 & 68.79 & 71.99 & 72.77 & \underline{73.47} & \textbf{73.56} & 73.26 \\
Swin-T & 81.24 & 73.09 & 63.85 & 72.18 & 73.78 & 64.12 & 71.71 & 73.38 & 75.92 & \underline{76.28} & \textbf{80.62} \\
SwinV2-S & 84.08 & 72.06 & 67.35 & 69.42 & 73.30 & 51.35 & 72.27 & 72.09 & 73.78 & \underline{73.87} & \textbf{74.91} \\
SwinV2-B & 84.62 & 72.43 & 66.72 & 71.06 & 74.87 & 51.91 & 73.37 & 72.24 & 74.34 & \underline{75.79} & \textbf{75.98} \\
SwinV2-L-In21k & 87.05 & 70.42 & 65.29 & 65.54 & 70.00 & 68.54 & 71.48 & 70.92 & 73.79 & \underline{74.87} & \textbf{76.48} \\
SwinV2-B-In21k & 86.64 & 70.21 & 64.45 & 65.17 & 69.36 & 66.65 & 71.82 & 71.30 & 73.96 & \underline{74.14} & \textbf{76.78} \\
ConvNeXt-T & 82.73 & 73.03 & 62.17 & 62.92 & 66.51 & 56.03 & 72.59 & 74.05 & \underline{75.05} & 75.05 & \textbf{77.07} \\
ConvNeXt-B & 84.41 & 73.52 & 58.05 & 63.96 & 70.50 & 56.61 & 74.05 & 73.67 & 75.22 & \underline{76.33} & \textbf{76.82} \\
ConvNeXt-B-In21k & 85.96 & 68.36 & 67.12 & 64.24 & 66.05 & 56.76 & 72.76 & 71.61 & \underline{73.94} & 73.93 & \textbf{75.83} \\
ConvNeXtV2-B-In21k & 87.09 & 66.43 & 66.17 & 64.84 & 65.40 & 48.34 & 71.04 & 69.60 & 71.68 & \underline{71.68} & \textbf{72.63} \\
ConvNeXtV2-T-In21k & 84.75 & 68.41 & 71.83 & 66.36 & 66.48 & 50.05 & 72.30 & 71.42 & \underline{72.88} & 72.88 & \textbf{74.15} \\
\midrule
\textbf{Average} & 84.31 & 69.46 & 66.12 & 67.99 & 69.02 & 57.66 & 72.04 & 71.72 & 73.51 & \underline{73.63} & \textbf{75.95} \\
\bottomrule
\end{tabular}%
}
\end{table*}

\subsection{Extended Results}
The extended results at Table \ref{tab:near_ood_shift_auroc} and Table \ref{tab:near_ood_shift_fpr_at_95}  show that MM++ achieves the best performance across different models under both AUROC and FPR95 metrics on Near-OOD benchmarks. MM++ achieves substantial improvements for models trained with standard supervised or augreg pipelines, where intermediate representations contains complementary semantic information that is effectively fused. The gains are particularly pronounced for architectures with strong hierarchical feature diversity, such as Swin and ConvNeXt. These models preserve complementary spatial and semantic statistics across layers. This behavior suggests that near-OOD shifts are pronounced at multiple abstraction levels rather than exclusively in the penultimate representation.

Across most DeiT3 variants, MM++ remains competitive but does not consistently outperform single-layer Mahalanobis++ baselines. 
DeiT3 models are trained with strong augmentations (e.g., Mixup, CutMix) and regularization techniques (e.g., label smoothing), which are known to produce smoother decision boundaries and more uniform feature representations.

As a result, intermediate layers tend to be more aligned with the final representation, reducing the diversity of information across layers. 
This limits the benefit of multi-layer fusion, as the representations become increasingly redundant, and single-layer methods already capture most of the discriminative structure.

Consequently, MM++ provides smaller gains in DeiT3 compared to other architectures, where stronger hierarchical feature diversity enables more effective cross-layer modeling.

\begin{table*}[htb]
\centering
\caption{FPR95 on Near-OOD averaged over ImageNet-V2, ImageNet-C, ImageNet-R, and ImageNet-ES with ImageNet-LT (ID)} 
\label{tab:near_ood_shift_fpr_at_95}

\renewcommand{\arraystretch}{1.3}
\resizebox{\textwidth}{!}{%
\begin{tabular}{>{\large}l >{\large}c >{\large}c >{\large}c >{\large}c >{\large}c >{\large}c >{\large}c >{\large}c >{\large}c >{\large}c >{\large}c}
\toprule
\large Model & \makecell{\normalsize Val\\Acc} & \normalsize MSP & \normalsize ODIN & \normalsize Energy & \normalsize ReAct & \normalsize KNN & \normalsize Maha & \normalsize rMaha & \normalsize Maha++ & \normalsize rMaha++ & \makecell{\normalsize MM++\\(ours)} \\

\midrule
ViT-B16-In21k-augreg & 84.47 & 73.95 & 72.49 & \underline{66.54} & 68.52 & 84.36 & 68.36 & 69.94 & 66.97 & 67.02 & \textbf{54.04} \\
ViT-L16-In21k-augreg & 85.81 & 74.56 & 67.79 & \underline{66.41} & 67.40 & 84.32 & 68.39 & 71.77 & 66.90 & 66.90 & \textbf{57.10} \\
ViT-T16-In21k-augreg & 75.48 & 70.38 & 71.96 & 60.06 & 62.91 & 67.40 & 82.10 & 80.69 & \underline{59.46} & 59.53 & \textbf{55.79} \\
ViT-S16-In21k-augreg & 81.37 & 72.08 & 70.28 & 63.27 & 64.28 & 79.68 & 74.78 & 74.75 & \underline{62.10} & 62.13 & \textbf{50.77} \\
ViT-B16-augreg & 79.13 & 73.45 & 80.12 & 70.13 & 70.40 & 90.37 & 71.41 & 73.19 & \underline{69.53} & 69.56 & \textbf{69.03} \\
ViT-S16-augreg & 78.79 & 73.10 & 79.24 & 69.50 & 68.95 & 95.26 & \underline{68.37} & 69.94 & 68.49 & 68.58 & \textbf{54.87} \\
ViT-so400M-SigLip & 89.33 & 81.48 & 82.42 & 78.19 & 76.79 & 93.18 & 73.66 & 74.39 & \textbf{69.49} & \underline{69.50} & 89.90 \\
ViT-B16-In21k-orig & 81.80 & 71.66 & 77.48 & \textbf{67.82} & \underline{68.05} & 90.86 & 71.11 & 70.76 & 68.51 & 68.32 & 69.20 \\
ViT-L16-In21k-orig & 81.50 & 72.44 & 78.91 & \underline{70.74} & 70.87 & 93.97 & 74.79 & 72.38 & \textbf{70.73} & 70.76 & 72.17 \\
ViT-B16-In21k-mil & 84.26 & 70.60 & 71.73 & \textbf{65.03} & \underline{65.21} & 91.83 & 83.64 & 77.92 & 68.20 & 67.63 & 74.67 \\
ViT-B16-In21k-augreg2 & 85.06 & 75.08 & 77.88 & 73.71 & 70.93 & 89.60 & 77.79 & 75.54 & 70.61 & \underline{70.51} & \textbf{59.25} \\
EVA02-L14-M38m-In21k & 89.56 & 81.88 & 83.30 & 81.63 & 81.73 & 94.98 & 78.63 & 78.55 & 77.33 & \underline{77.33} & \textbf{59.59} \\
EVA02-B14-In21k & 88.38 & 77.75 & 82.55 & 78.48 & 78.22 & 96.00 & 74.38 & 74.49 & \underline{72.11} & 72.11 & \textbf{56.89} \\
EVA02-S14 & 85.64 & 73.79 & 80.49 & 73.98 & 74.04 & 85.13 & 68.99 & 69.45 & 67.04 & \underline{67.04} & \textbf{54.59} \\
EVA02-T14 & 80.58 & 73.74 & 81.42 & 71.62 & 71.88 & 81.21 & 70.05 & 70.51 & 66.62 & \underline{66.62} & \textbf{65.29} \\
DeiT3-B16 & 83.72 & 75.69 & 77.51 & 79.73 & 81.84 & 80.43 & 77.76 & 76.56 & \underline{75.48} & \textbf{75.13} & 79.00 \\
DeiT3-L16-In21k & 87.58 & 82.17 & 81.54 & 79.13 & 75.61 & 83.66 & 74.35 & 74.12 & \textbf{73.71} & \underline{73.75} & 74.19 \\
DeiT3-L16 & 85.80 & 78.52 & 79.38 & 85.53 & 92.13 & 71.16 & 71.05 & 71.19 & \textbf{68.91} & \underline{69.61} & 72.14 \\
DeiT3-B16-In1k & 84.95 & 75.24 & 80.29 & 82.92 & 86.47 & 76.84 & 74.14 & 73.85 & \underline{72.79} & \textbf{72.48} & 74.97 \\
DeiT3-B16-384-In1k & 86.67 & 79.50 & 82.45 & 78.94 & 74.55 & 94.66 & 72.14 & 71.45 & \textbf{70.45} & \underline{70.46} & 72.26 \\
DeiT3-B16-224-In1k & 85.72 & 79.17 & 81.61 & 78.24 & 74.34 & 94.05 & 74.04 & 72.70 & \underline{71.62} & \textbf{71.61} & 73.90 \\
DeiT3-S16-In21k & 84.74 & 76.95 & 82.12 & 74.59 & 72.42 & 97.33 & 73.50 & 72.81 & \textbf{70.30} & \underline{70.30} & 73.47 \\
DeiT3-S16 & 83.44 & 72.79 & 76.89 & 76.67 & 77.99 & 76.02 & 76.46 & 75.15 & \underline{70.60} & \textbf{70.46} & 79.77 \\
Swin-T & 81.24 & 71.82 & 84.11 & 70.40 & 68.37 & 78.86 & 77.32 & 74.19 & 64.63 & \underline{64.31} & \textbf{54.29} \\
SwinV2-S & 84.08 & 72.20 & 78.92 & 71.98 & 67.90 & 91.94 & 73.56 & 72.72 & 66.71 & \underline{66.64} & \textbf{65.52} \\
SwinV2-B & 84.62 & 72.32 & 76.94 & 68.03 & \textbf{63.71} & 88.13 & 72.01 & 71.30 & 67.28 & 66.25 & \underline{66.18} \\
SwinV2-L-In21k & 87.05 & 75.33 & 81.53 & 79.36 & 77.94 & 72.17 & 74.94 & 73.02 & 67.63 & \underline{66.94} & \textbf{64.36} \\
SwinV2-B-In21k & 86.64 & 74.49 & 80.91 & 76.13 & 73.38 & 73.32 & 73.79 & 72.54 & 67.14 & \underline{67.09} & \textbf{64.48} \\
ConvNeXt-T & 82.73 & \underline{70.34} & 85.46 & 84.54 & 82.96 & 91.87 & 75.19 & 73.09 & 71.31 & 71.37 & \textbf{66.30} \\
ConvNeXt-B & 84.41 & \underline{68.98} & 85.34 & 82.26 & 76.88 & 83.26 & 72.79 & 71.75 & 71.60 & 73.80 & \textbf{68.65} \\
ConvNeXt-B-In21k & 85.96 & 74.43 & 76.18 & 79.38 & 76.65 & 92.18 & 68.40 & 69.35 & \underline{66.66} & 66.66 & \textbf{64.28} \\
ConvNeXtV2-B-In21k & 87.09 & 76.87 & 79.85 & 75.50 & 75.20 & 95.01 & 71.20 & 71.49 & \underline{70.18} & 70.18 & \textbf{69.40} \\
ConvNeXtV2-T-In21k & 84.75 & 73.65 & 72.06 & 74.51 & 74.45 & 95.29 & 68.87 & 69.79 & \underline{67.30} & 67.30 & \textbf{65.96} \\
\midrule
\textbf{Average} & 84.31 & 74.74 & 78.82 & 74.39 & 73.73 & 86.50 & 73.57 & 73.07 & 69.04 & \underline{69.03} & \textbf{66.43} \\
\bottomrule
\end{tabular}%
}
\end{table*}

\paragraph{Summary.}
Across diverse architectures—including standard transformers, hierarchical transformers, and convolutional networks—MM++ consistently improves performance on near-OOD benchmarks. While single-layer methods remain effective for large semantic shifts, they fail to capture subtle distribution changes.

In contrast, MM++ leverages entropy-guided layer selection and joint multi-layer modeling to detect these shifts more effectively. Importantly, this is achieved in a strictly post-hoc setting without fine-tuning, making MM++ both practical and broadly applicable across heterogeneous architectures.

\begin{table}[htbp]
\centering
\caption{Ablation of covariance estimation and calibration-set size for Maha++ on ViT-B/16 with ImageNet-1K as ID. Values are AUROC (\%). EC denotes empirical covariance and LW denotes Ledoit--Wolf shrinkage.}
\label{tab:maha_cov_ablation}
\vspace{0.2cm}
\resizebox{0.9\linewidth}{!}{%
\begin{tabular}{lccccc}
\toprule
\textbf{Method} & \textbf{iNaturalist} & \textbf{SUN} & \textbf{Places365} & \textbf{Textures} & \textbf{ImageNet-O} \\
\midrule
Maha++ EC (full 1.28M) & \underline{98.76} & \textbf{88.90} & 86.48 & 89.26 & \underline{86.42} \\
Maha++ LW (full 1.28M) & \textbf{98.79} & \textbf{88.90} & \underline{86.49} & \underline{89.27} & \textbf{86.44} \\
Maha++ LW (115K)       & \underline{98.76} & \underline{88.88} & \textbf{86.54} & \textbf{89.28} & 86.39 \\
Maha++ EC (115K)       & 98.74 & \underline{88.88} & \textbf{86.54} & \textbf{89.28} & 86.38 \\
\bottomrule
\end{tabular}
}
\end{table}








\subsection{Additional Ablation Results}
\noindent\textbf{Covariance estimation ablation.}
Table~\ref{tab:maha_cov_ablation} evaluates the impact of covariance estimation methods and calibration-set size on the single-layer Mahalanobis++ baseline. We compare empirical covariance (EC) and Ledoit--Wolf (LW) shrinkage using both the full ImageNet-1K training set (1.28M samples) and a reduced subset of 115K samples. Across all five OOD benchmarks, the performance differences between EC and LW are marginal, with AUROC variations remaining within 0.05\%. Similarly, reducing the calibration set size from 1.28M to 115K samples has a negligible impact. These findings suggest that, following $\ell_2$ normalization, the single penultimate-layer feature distribution is already sufficiently well-conditioned; thus, standard EC estimation remains highly stable and data-efficient. Consequently, we retain EC for the standard Mahalanobis++ baseline in our experiments to maintain consistency with the original protocol~\cite{mueller2025mahalanobisPP}.

Furthermore, these results demonstrate that applying LW shrinkage in isolation—without top-$K$ information gating and feature fusion—yields limited performance gains. This observation explicitly motivates the design of our proposed framework. While a single layer is inherently well-conditioned, MM++ concatenates the top $K-1$ intermediate layers with the penultimate layer into a unified, high-dimensional representation space. Because this architecture-independent feature fusion creates a complex joint distribution prone to ill-conditioning, MM++ crucially relies on LW shrinkage to regularize the precision matrix, allowing the model to effectively leverage multi-layer information for robust OOD detection.




\begin{table}[htbp]
\centering
\caption{Practical overhead comparison on ViT-B/16}
\label{tab:overhead}
\vspace{0.2cm}
\resizebox{0.95\textwidth}{!}{%
\begin{tabular}{l ccc cc}
\toprule
&
\multicolumn{2}{c}{\textbf{Offline}} &
\multicolumn{3}{c}{\textbf{Online}} \\
\cmidrule(lr){2-3} \cmidrule(lr){4-6} 

\textbf{Method}
& \textbf{Fine-tuning}
& \textbf{Total Time}
& \textbf{\# of Layers}
& \textbf{Memory/sample}
& \textbf{GPU Latency} \\

\midrule
Maha++       & None    & $\sim$16s    & 1  & 3 KB  & 0.0009 ms \\
MM++ ($K=2$) & None    & $\sim$55s    & 2  & 6 KB  & 0.0020 ms \\
X-Maha       & $\sim$29h  & $\sim$29h       & 12 & 36 KB & 0.0464 ms \\
\bottomrule
\end{tabular}
}
\end{table}

\begin{figure}[htb]
      \centering
    \resizebox{\textwidth}{!}{%
      \begin{subfigure}[b]{0.34\linewidth}
        \centering
        \includegraphics[width=\linewidth]{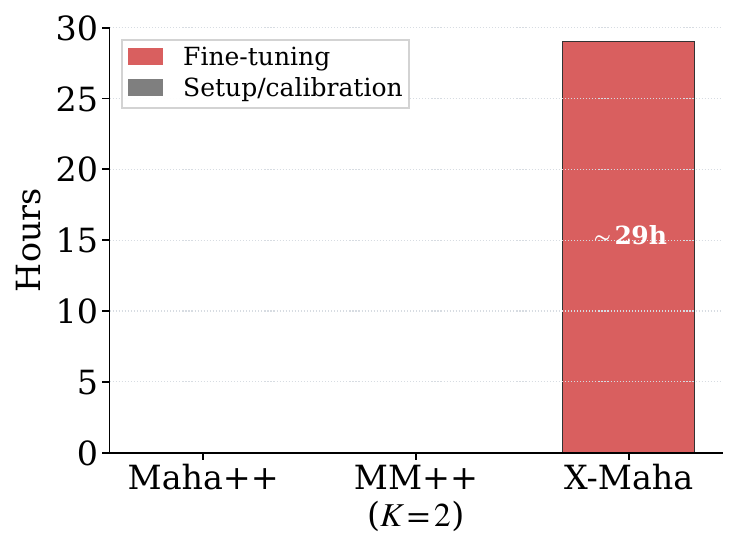}
        \caption{Offline}
        \label{fig:overhead-offline}
      \end{subfigure}
      \hfill
      \begin{subfigure}[b]{0.66\linewidth}
        \centering
        \includegraphics[width=\linewidth]{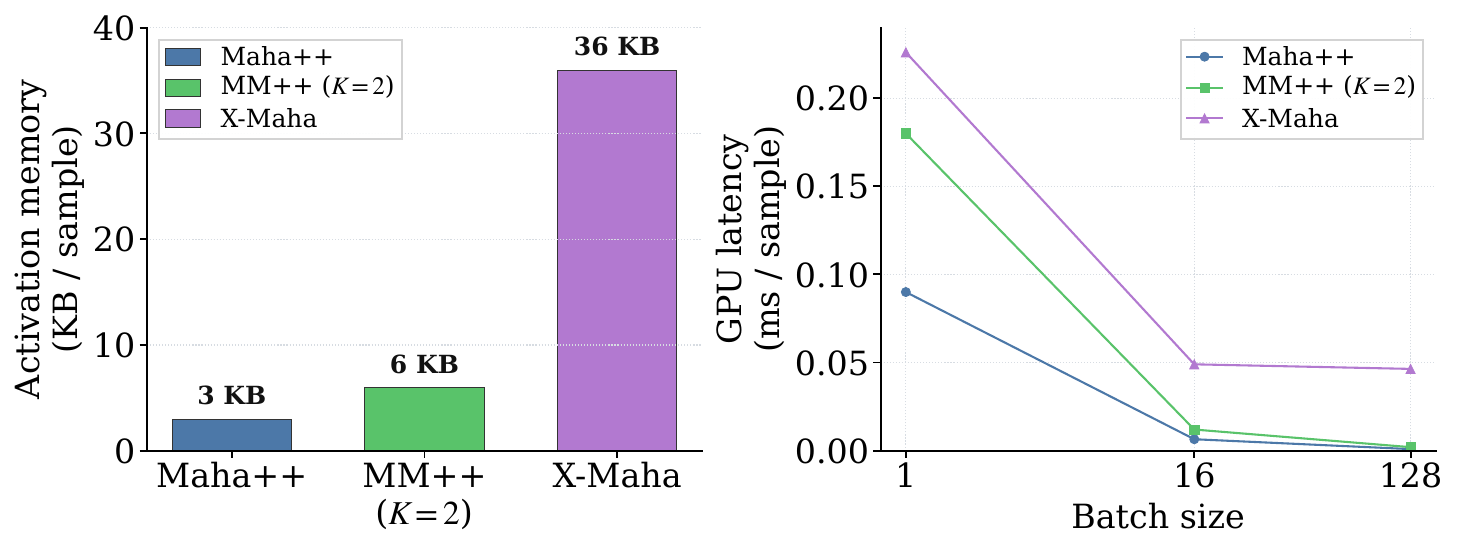}
        \caption{Online}
        \label{fig:overhead-online}
      \end{subfigure}
      }
      \caption{Practical overhead comparison on ViT-B/16.}
      \label{fig:overhead-vitb16}
\end{figure}

\noindent\textbf{Efficiency Ablation.}
We compare the deployment cost of MM++ with Mahalanobis++ and X-Mahalanobis in Table~\ref{tab:overhead} and Fig.~\ref{fig:overhead-vitb16}.
As shown 
in Fig.\ref{fig:overhead-offline},
X-Mahalanobis incurs substantial offline overhead due to fine-tuning (${\sim}29$h) particularly when ImageNet-1K full training set is used as the ID dataset.
In contrast, both Mahalanobis++ and MM++ require no fine-tuning and remain fully post-hoc with lightweight calibration (16\,s and 55\,s, respectively).

For online OOD detection, MM++ uses only $K{=}2$ layers, compared to 12 layers in X-Mahalanobis, resulting in significantly lower memory usage and latency.
As illustrated in Fig.~\ref{fig:overhead-online},
MM++ requires only 6\,KB of activation memory and achieves $0.002$\,ms GPU latency per sample, remaining close to Mahalanobis++ while being substantially more efficient than X-Mahalanobis (36\,KB, $0.046$\,ms).
Overall, MM++ preserves the efficiency of post-hoc methods while enabling multi-layer modeling without expensive fine-tuning.

\section{Reproduction Details}
\label{app:reproduce}
\input{App-reproduce}

\section{Methods}
\label{app:methods}
\input{App-methods}

\input{App-limitations}
\input{App-broad}

\input{App-Licenses}

%% file: App-justify3.tex
\section{Theoretical Justification of MM++}
\label{app:justify}

The effectiveness of MM++ for out-of-distribution (OOD) detection is supported by three complementary principles: (i) entropy-based layer selection, (ii) joint precision modeling across layers, and (iii) stable covariance estimation via shrinkage in high-dimensional fused spaces.

\paragraph{1. Entropy-Based Layer Selection and Neural Collapse.}
The selection of informative layers via the entropy density drop $\Delta_l$ (Eq.~\ref{eq:delta}) is motivated by the geometric properties of neural collapse \cite{papyan2020prevalence}. As representations propagate through a deep network, within-class variability transitions from high-dimensional, distributed features in early layers to low-dimensional, highly structured embeddings in later layers. 

Because our representations are $\ell_2$-normalized onto the unit hypersphere, the within-class covariance entropy $H_l$ acts as a direct proxy for the active intrinsic dimensionality of the ID data manifold. A sharp increase in $\Delta_l$ identifies the precise network depth where non-discriminative variability is aggressively suppressed, projecting ID samples onto a tightly constrained subspace. By selecting layers with maximal $\Delta_l$, MM++ isolates the boundaries of strongest semantic compression. OOD samples, lacking the semantic structure of ID classes, are not projected onto this same low-dimensional manifold, resulting in a measurable orthogonal deviation that the Mahalanobis distance subsequently penalizes.

\paragraph{2. Joint Precision and Cross-Layer Consistency.}
Multi-layer OOD detectors such as \cite{xmahalanobis2025, lee2018simple} typically rely on additive fusion, which implicitly assumes independence across layers. This corresponds to approximating the joint covariance as a block-diagonal matrix, i.e., $\Sigma_{ll'} = \mathbf{0}$ for $l \neq l'$, thereby ignoring cross-layer dependencies.

Unlike methods that compute independent OOD scores for each layer and aggregate them post-hoc, \textbf{MM++ computes a single Mahalanobis++ distance in a unified representation space.} We estimate a joint precision matrix $\hat{\Sigma}_{\mathcal{K}}^{-1}$ across the entire set of selected layers $\mathcal{K}$. By defining $\phi_c(x)$ as the concatenated, $\ell_2$-normalized feature residual vector across all $l \in \mathcal{K}$, the class-conditional score is expressed as a single quadratic form:
\begin{equation}
    \mathcal{S}_c(x) = - \phi_c(x)^\top \hat{\Sigma}_{\mathcal{K}}^{-1} \phi_c(x).
\end{equation}

To illustrate why this differs fundamentally from a layer-wise calculation, we partition $\hat{\Sigma}_{\mathcal{K}}^{-1}$ into block components. The score expands as:
\begin{equation}
    \mathcal{S}_c(x) = \underbrace{\sum_{l \in \mathcal{K}} \mathcal{S}_{c,l}(x)}_{\text{Intra-layer terms}} + \underbrace{\sum_{l \in \mathcal{K}} \sum_{l' \neq l} \mathcal{S}_c^{(l,l')}(x)}_{\text{Inter-layer interactions}},
    \label{eq:inter}
\end{equation}
where $\mathcal{S}_{c,l}(x) = -\phi_{c,l}(x)^\top \hat{\Sigma}^{-1}_{\mathcal{K},ll} \phi_{c,l}(x)$ evaluates the marginal fit within layer $l$. Crucially, the cross-layer term $\mathcal{S}_c^{(l,l')}(x)$ is governed by the off-diagonal blocks:
\begin{equation}
    \mathcal{S}_c^{(l,l')}(x) = - \phi_{c,l}(x)^\top \hat{\Sigma}_{\mathcal{K}, ll'}^{-1} \phi_{c,l'}(x).
\end{equation}

Mathematically, these off-diagonal precision blocks act as conditional expectation penalties. Under a joint Gaussian assumption, the optimal prediction of a feature at layer $l'$ given the feature at layer $l$ depends heavily on their cross-covariance. The second term in Eq.~\ref{eq:inter} explicitly penalizes deviations from the learned ID feature trajectory $\mathbb{E}[\phi_{c,l'} | \phi_{c,l}]$. Consequently, MM++ detects sophisticated OOD samples that might perfectly match the marginal statistics of layer $l$ and layer $l'$ independently, but violate the expected cross-layer evolutionary dynamics dictated by the ID data.

\paragraph{3. Well-Conditioned Estimation via Ledoit--Wolf Shrinkage.}
While concatenated fusion unlocks cross-layer modeling, it significantly exacerbates the dimensionality problem. Let $D_{\mathcal{K}} = \sum_{l \in \mathcal{K}} D_l$ be the dimension of the joint space. When $D_{\mathcal{K}} \gg N_c$, the empirical joint covariance $\hat{\Sigma}_{\mathcal{K}}$ becomes highly ill-conditioned or strictly rank-deficient (possessing zero-valued eigenvalues), meaning the raw inverse $\hat{\Sigma}_{\mathcal{K}}^{-1}$ does not exist. Even when invertible, the empirical estimator systematically overestimates dominant eigenvalues and severely underestimates small eigenvalues. In the Mahalanobis distance computation, directions corresponding to these underestimated small eigenvalues are inverted, yielding an unstable, near-infinite penalty for negligible noise.

To guarantee a theoretically stable and positive-definite precision matrix, MM++ applies Ledoit--Wolf shrinkage (Eq.~\ref{eq:shrinkage}) directly to the fused space:
\begin{equation}
    \hat{\Sigma}_{\mathcal{K},\text{shrink}} = (1 - \gamma) \hat{\Sigma}_{\mathcal{K}} + \gamma\,\left(\frac{\operatorname{Tr}(\hat{\Sigma}_{\mathcal{K}})}{D_{\mathcal{K}}}\right)\mathbf{I},
\end{equation}
where $\gamma \in (0,1)$ is the analytically optimal shrinkage intensity. 

This operation rigidly bounds the eigenvalue spectrum of the resulting precision matrix. Let $\lambda_i \ge 0$ be an eigenvalue of the empirical $\hat{\Sigma}_{\mathcal{K}}$. The corresponding eigenvalue of the precision matrix $\hat{\Sigma}_{\mathcal{K},\text{shrink}}^{-1}$ becomes:
\begin{equation}
    \tilde{\lambda}_i^{-1} = \frac{1}{(1-\gamma)\lambda_i + \gamma \frac{\operatorname{Tr}(\hat{\Sigma}_{\mathcal{K}})}{D_{\mathcal{K}}}}.
\end{equation}
As $\lambda_i \to 0$, the precision eigenvalue is safely bounded by a maximum theoretical value of $D_{\mathcal{K}} / (\gamma \operatorname{Tr}(\hat{\Sigma}_{\mathcal{K}}))$. Thus, shrinkage acts as an optimal, data-driven Tikhonov regularizer. It suppresses spurious responses to high-frequency noise in the concatenated null-space, ensuring that the OOD score $\mathcal{S}_c(x)$ is driven by meaningful semantic deviations rather than numerical instability.

\paragraph{Summary.}
Entropy-based layer selection identifies the geometric boundaries of semantic compression, joint precision modeling captures conditional cross-layer dependencies, and analytically derived shrinkage ensures the necessary topological stability of the high-dimensional fused space. Together, these components mathematically underpin the improved separability of MM++.

%% file: App-reproduce.tex


\textbf{Evaluation Rigor and Stability.} Because MM++ is a strictly post-hoc framework operating on frozen, standardized model checkpoints, the extracted feature representations and resulting OOD scores are algorithmically deterministic. 

Consequently, we empirically validate the robustness of our method across multiple distinct architectural paradigms and diverse distribution shifts, rather than through stochastic repetition. 
The consistent performance of MM++ across multiple, distinct architectural designs (global attention, hierarchical attention, and modernized convolutions) and under highly varied distribution shifts (from long-tailed class imbalance to severe covariate degradation) serves as a comprehensive validation of its algorithmic robustness.
 
\textbf{Hyperparameter, Feature Extraction, and Well-Conditioned Covariance Estimation.} MM++ is a data-driven, training-free method with a single hyperparameter ($K=2$). We use publicly available pretrained backbones (e.g., ViT-B/16, Swin-T, and ConvNeXt-T) and standard OOD benchmarks with their official splits. Features are extracted from intermediate layers, $\ell_2$-normalized, and class-conditional statistics are computed from in-distribution training data.

The joint covariance matrix over the selected layers is estimated using Ledoit--Wolf shrinkage, and OOD scores are computed via the Mahalanobis++ formulation described in Section~\ref{sec:method}. No additional training or hyperparameter tuning is required. Moreover, our detailed experimental set-up is described in Section~\ref{sec:eval}.

\textbf{Code Release.} An anonymized ZIP archive containing the complete implementation is included with the submission. Upon acceptance, the codebase will be made publicly available. Our implementation ensures full reproducibility of all reported results using the publicly accessible OOD benchmarks detailed in this work.

\textbf{Compute Resources.}
All our calibration and evaluation are performed using on a workstation with an Intel Core i9 processor (16 cores), 32 GB DDR5 SDRAM, and an NVIDIA RTX 4090 GPU with 24GB GDDR6X SDRAM.



%% file: App-methods.tex

We briefly review the OOD detection methods evaluated in this paper.

Let $f(x) \in \mathbb{R}^C$ denote the pre-softmax logits of the network, and $h(x) \in \mathbb{R}^d$ denote the penultimate feature representation (with $h_l(x)$ for layer-wise features).

\begin{itemize}
    \item \textbf{MSP \cite{hendrycks2017baseline}:} Uses the maximum softmax probability as the confidence score:
    \[
        \mathcal{S}_{\text{MSP}}(x) = \max_c \frac{\exp(f_c(x))}{\sum_{j} \exp(f_j(x))}
    \]

    \item \textbf{ODIN \cite{liang2018enhancing}:} Applies temperature scaling ($T$) and input perturbation ($x \rightarrow \tilde{x}$):
    \[
        \mathcal{S}_{\text{ODIN}}(x) = \max_c \frac{\exp(f_c(\tilde{x})/T)}{\sum_{j} \exp(f_j(\tilde{x})/T)}
    \]

    \item \textbf{Energy \cite{liu2020energy}:} Uses the free energy of logits as an unnormalized confidence score:
    \[
        \mathcal{S}_{\text{Energy}}(x) = T \log \sum_{c} \exp(f_c(x)/T)
    \]

    \item \textbf{ReAct \cite{sun2021react}:} Truncates the penultimate activations $h(x)$ at a threshold $v$ to produce rectified logits $f^{\text{clip}}(x)$, computing the energy score as:
    \[
        \mathcal{S}_{\text{ReAct}}(x) = T \log \sum_{c=1}^{C} \exp(f_c^{\text{clip}}(x)/T)
    \]  

    \item \textbf{KNN \cite{sun2022knn}:} Uses the negative distance to the $k$-th nearest neighbor in feature space:
    \[
        \mathcal{S}_{\text{KNN}}(x) = - \| h(x) - h_{(k)}(x) \|_2
    \]

    \item \textbf{Mahalanobis \cite{lee2018simple}:} Models features as class-conditional Gaussians with shared covariance $\Sigma$:
    \[
        \mathcal{S}_{\text{Maha}}(x) = - \min_c (h(x) - \mu_c)^\top \Sigma^{-1} (h(x) - \mu_c)
    \]

    \item \textbf{Relative Mahalanobis \cite{ren2021simple}:} Refines the Mahalanobis score by subtracting a background distance term based on $(\mu_0, \Sigma_0)$:
    \[
        \mathcal{S}_{\text{rMaha}}(x) = - \min_c \left[ d_c(x) - d_0(x) \right],
        \text{where } 
    \]
    \[
        d_c(x) = (h(x) - \mu_c)^\top \Sigma^{-1} (h(x) - \mu_c), \quad
        d_0(x) = (h(x) - \mu_0)^\top \Sigma_0^{-1} (h(x) - \mu_0).
    \]

    \item \textbf{Relative Mahalanobis++:} Extends relative Mahalanobis by using $\ell_2$-normalized features $\tilde{h}(x)$:
    \[
        \mathcal{S}_{\text{rMaha++}}(x) = - \min_c \Big[ (\tilde{h}(x) - \tilde{\mu}_c)^\top \tilde{\Sigma}^{-1} (\tilde{h}(x) - \tilde{\mu}_c) 
        - (\tilde{h}(x) - \tilde{\mu}_0)^\top \Sigma_0^{-1} (\tilde{h}(x) - \tilde{\mu}_0) \Big]
    \]

    \item \textbf{X-Mahalanobis \cite{xmahalanobis2025}:} Linearly aggregates layer-wise Mahalanobis distances using variance-based weights $\alpha_l$:
    \[
        \mathcal{S}_{\text{X-Maha}}(x) = - \min_{c} \sum_{l=1}^{L} \alpha_l (h_l(x) - \mu_{c,l})^\top \Sigma_l^{-1} (h_l(x) - \mu_{c,l})
    \]

    \item \textbf{Mahalanobis++ \cite{mueller2025mahalanobisPP}:} Applies a scale-invariant Mahalanobis distance on $\ell_2$-normalized features:
    \[
        \mathcal{S}_{\text{Maha++}}(x) = - \min_c ( \tilde{h}(x) - \tilde{\mu}_c )^\top \tilde{\Sigma}^{-1} ( \tilde{h}(x) - \tilde{\mu}_c )
    \]

    \item \textbf{MM++ (Ours):} Evaluates a single joint Mahalanobis++ distance on a unified representation space $\phi_c(x)$, fusing $\ell_2$-normalized features from the terminal (penultimate) layer and the top $K-1$ intermediate layers. The tied precision matrix $\hat{\Sigma}_{\mathcal{K}}^{-1}$ is derived analytically via Ledoit--Wolf shrinkage:
    \[
        \mathcal{S}_{\text{MM++}}(x) = -\min_{c} \bigl(\boldsymbol{\phi}(x) - \hat{\boldsymbol{\mu}}_c^{\mathcal{K}}\bigr)^{\!\top} \hat{\Sigma}_{\mathcal{K}}^{-1} \bigl(\boldsymbol{\phi}(x) - \hat{\boldsymbol{\mu}}_c^{\mathcal{K}}\bigr)
    \]
\end{itemize}

%% file: App-limitations.tex
\section{Limitations}
\label{app:limitations}

MM++ incurs additional computational and memory overhead compared to single-layer methods, due to eigenspectrum estimation of within-class covariance matrices and storage of layer-wise statistics. However, the overhead is minimal particularly when $K$ is small (e.g., $K = 2$) as shown in Table~\ref{tab:overhead}. This is because Top-$K$ information gating selects $K-1$ intermediates layers with the sharpest entropy density drops, while anchoring the penultimate layer.

As a strictly post-hoc method, MM++ also depends on the geometry of the pretrained representation space. Its gating mechanism assumes identifiable entropy density drops ($\Delta_l$) associated with semantic compression; when such structure is weak or absent, intermediate features may offer limited discriminative signal for near-OOD detection. For this reason, MM++ does not select intermediate layer $l$ if $\Delta_l \le 0$.

Extending MM++ to dense prediction tasks remains an open challenge, requiring spatially localized estimates of intrinsic dimensionality to account for heterogeneous feature collapse.





%% file: App-broad.tex
\section{Broader Impacts}
\label{sec:broader_impacts}

This work improves the reliability of deep neural networks in open-world settings by enabling unsupervised, post-hoc out-of-distribution (OOD) detection. The proposed approach reduces computational overhead by avoiding retraining and does not rely on labeled proxy OOD data, thereby limiting potential sources of bias.

However, MM++ shares limitations common to OOD detection methods. It is not guaranteed to detect all anomalous inputs, and over-reliance on its outputs may lead to insufficient oversight. As such, it should be used as part of a broader safety framework that includes complementary safeguards, such as uncertainty estimation or human supervision.




%% file: App-Licenses.tex
\section{Public Resources Used}
\label{app:resources}
In this section, we acknowledge the public resources used during the course of this work.

\subsection{Public Datasets Used}

\begin{itemize}

    \item \textbf{OpenOOD} \dotfill MIT License
    \item \textbf{NINCO} \dotfill MIT License
    \item \textbf{ImageNet-1K} \dotfill ImageNet Agreement
    \item \textbf{ImageNet-LT} \dotfill BSD-3-Clause license
    \item \textbf{Texture} \dotfill Creative Commons BY-SA 4.0
    \item \textbf{OpenImage-O} \dotfill Apache License 2.0
    \item \textbf{ImageNet-C} \dotfill ImageNet Agreement
    \item \textbf{ImageNet-ES} \dotfill ImageNet Agreement
    \item \textbf{ImageNet-R} \dotfill ImageNet Agreement
    \item \textbf{ImageNet-V2} \dotfill ImageNet Agreement
    \item \textbf{Places365} \dotfill Creative Commons Attribution 4.0 License
    \item \textbf{iNaturalist} \dotfill Mixed Creative Commons Licenses
     \item \textbf{SUN} \dotfill SUN Database License

\end{itemize}

\subsection{Public Implementations Used}

\begin{itemize}

    \item \textbf{Mahalanobis++} \dotfill Creative Commons BY 4.0 License
    \item \textbf{timm} \dotfill Apache License 2.0
    \item \textbf{scikit-learn} \dotfill BSD-3-Clause License 
    \item \textbf{X-Mahalanobis} \dotfill Creative Commons BY 4.0 License
    \item \textbf{ODIN} \dotfill MIT License
    \item \textbf{Energy} \dotfill MIT License
    \item \textbf{ReAct} \dotfill MIT License

\end{itemize}


%% file: reference.bib
@inproceedings{
du2022towards,
title={VOS: Learning What You Don't Know by Virtual Outlier Synthesis},
author={Xuefeng Du and Zhaoning Wang and Mu Cai and Sharon Li},
booktitle={International Conference on Learning Representations},
year={2022},
url={https://openreview.net/forum?id=TW7d65uYu5M}
}

@article{shwartz2017opening,
  title={Opening the black box of deep neural networks via information},
  author={Shwartz-Ziv, Ravid and Tishby, Naftali},
  journal={arXiv preprint arXiv:1703.00810},
  year={2017}
}

@inproceedings{martins2016softmax,
  title={{From softmax to sparsemax: A sparse model of attention and multi-label classification}},
  author={Martins, Andr{\'e} FT and Astudillo, Ram{\'o}n Fernandez},
  booktitle={International Conference on Machine Learning},
  pages={1614--1623},
  year={2016},
  organization={PMLR}
}

@article{yang2022openood,
  title={Openood: Benchmarking generalized out-of-distribution detection},
  author={Yang, Jingkang and Wang, Pengyun and Zou, Dejian and Zhou, Zitang and Ding, Kunyuan and Peng, Wenxuan and Wang, Haoqi and Chen, Guangyao and Li, Bo and Sun, Yiyou and others},
  journal={Advances in Neural Information Processing Systems},
  volume={35},
  pages={32598--32611},
  year={2022}
}

@article{ledoit2004well,
  title={A well-conditioned estimator for large-dimensional covariance matrices},
  author={Ledoit, Olivier and Wolf, Michael},
  journal={Journal of multivariate analysis},
  volume={88},
  number={2},
  pages={365--411},
  year={2004},
  publisher={Elsevier}
}

@inproceedings{tishby2015deep,
  title={Deep learning and the information bottleneck principle},
  author={Tishby, Naftali and Zaslavsky, Noga},
  booktitle={Proceedings of the IEEE Information Theory Workshop},
  year={2015}
}

@article{harun2025controlling,
  title={Controlling Neural Collapse Enhances Out-of-Distribution Detection and Transfer Learning},
  author={Harun, Md Yousuf and Gallardo, Jhair and Kanan, Christopher},
  journal={arXiv preprint arXiv:2502.10691},
  year={2025}
}

@article{janiak2025geometry,
  title={A Geometry-Based View of Mahalanobis {OOD} Detection},
  author={Denis Janiak and Jakub Binkowski and Tomasz Kajdanowicz}, 
  journal={arXiv preprint arXiv:2510.15202},
  year={2025}
}

@inproceedings{neco2024,
  title={{NECO}: NEural Collapse Based Out-of-distribution detection},
  author={Ammar, Mou{\"\i}n Ben and Belkhir, Nacim and Popescu, Sebastian and Manzanera, Antoine and Franchi, Gianni},
  booktitle={Proceedings of the International Conference on Learning Representations},
  year={2024}
}

@inproceedings{lin2021mood,
  title={Mood: Multi-level out-of-distribution detection},
  author={Lin, Ziqian and Roy, Suman and Li, Yixuan},
  booktitle={Proceedings of the IEEE/CVF Conference on Computer Vision and Pattern Recognition (CVPR)},
  pages={15313--15323},
  year={2021}
}

@article{papyan2020prevalence,
  title={Prevalence of neural collapse during the terminal phase of deep learning training},
  author={Papyan, Vardan and Han, XY and Donoho, David L},
  journal={Proceedings of the National Academy of Sciences},
  volume={117},
  number={40},
  pages={24652--24663},
  year={2020},
  publisher={National Academy of Sciences}
}

@article{lu2025out,
  title={{Out-of-distribution detection: A task-oriented survey of recent advances}},
  author={Lu, Shuo and Wang, Yingsheng and Sheng, Lijun and He, Lingxiao and Zheng, Aihua and Liang, Jian},
  journal={ACM Computing Surveys},
  volume={58},
  number={2},
  pages={1--39},
  year={2025},
  publisher={ACM New York, NY}
}

@article{yang2024generalized,
  title={{Generalized out-of-distribution detection: A survey}},
  author={Yang, Jingkang and Zhou, Kaiyang and Li, Yixuan and Liu, Ziwei},
  journal={International Journal of Computer Vision},
  volume={132},
  number={12},
  pages={5635--5662},
  year={2024},
  publisher={Springer}
}

@inproceedings{regmi2024t2fnorm,
  title={{T2fnorm: Train-time feature normalization for OOD detection in image classification}},
  author={Regmi, Sudarshan and Panthi, Bibek and Dotel, Sakar and Gyawali, Prashnna K and Stoyanov, Danail and Bhattarai, Binod},
  booktitle={Proceedings of the IEEE/CVF Conference on Computer Vision and Pattern Recognition},
  pages={153--162},
  year={2024}
}

@article{Haas2023ExploringSH,
  title={Exploring Simple, High Quality Out-of-Distribution Detection with L2 Normalization},
  author={Jarrod Haas and William Yolland and Bernhard T. Rabus},
  journal={Transactions on Machine Learning Research},
  volume = {2024},
  year={2024}
}

@inproceedings{ming2022exploit,
    title={How to exploit hyperspherical embeddings for out-of-distribution detection?},
    author={Ming, Yifei and Sun, Yiyou and Dia, Ousmane and Li, Yixuan},
    booktitle = {Proceedings of the International Conference on Learning Representations},
    year = 2023
}

@inproceedings{sehwag2021ssd,
  title={{SSD: A unified framework for self-supervised outlier detection}},
  author={Sehwag, Vikash and Chiang, Mung and Mittal, Prateek},
  booktitle={Proceedings of the International Conference on Learning Representations},
  year={2021}
}

@inproceedings{cimpoi2014describing,
  title={Describing textures in the wild},
  author={Cimpoi, Mircea and Maji, Subhransu and Kokkinos, Iasonas and Vedaldi, Andrea and Soatto, Stefano},
  booktitle={Proceedings of the IEEE conference on computer vision and pattern recognition},
  pages={3606--3613},
  year={2014}
}

@article{zhou2017places,
  title={Places: A 10 million image database for scene recognition},
  author={Zhou, Bolei and Lapedriza, Agata and Khosla, Aditya and Oliva, Aude and Torralba, Antonio},
  journal={IEEE Transactions on Pattern Analysis and Machine Intelligence},
  volume={40},
  number={6},
  pages={1452--1464},
  year={2017},
  publisher={IEEE}
}

@inproceedings{bitterwolf2023ninco,
  title={{In or out? Fixing ImageNet out-of-distribution detection evaluation}},
  author={Bitterwolf, Julian and M{\"u}ller, Maximilian and Hein, Matthias},
  booktitle={Proceedings of the International Conference on Machine Learning},
  pages={2441--2472},
  year={2023}
}

@inproceedings{hendrycks2021natural,
  title={Natural adversarial examples},
  author={Hendrycks, Dan and Zhao, Kevin and Basart, Steven and Steinhardt, Jacob and Song, Dawn},
  booktitle={Proceedings of the IEEE/CVF Conference on Computer Vision and Pattern Recognition},
  pages={15262--15271},
  year={2021}
}

@article{russakovsky2015imagenet,
  title={Imagenet large scale visual recognition challenge},
  author={Russakovsky, Olga and Deng, Jia and Su, Hao and Krause, Jonathan and Satheesh, Sanjeev and Ma, Sean and Huang, Zhiheng and Karpathy, Andrej and Khosla, Aditya and Bernstein, Michael and Berg, Alexander C. and Fei-Fei, Li},
  journal={International Journal of Computer Vision},
  volume={115},
  number={3},
  pages={211--252},
  year={2015},
  publisher={Springer}
}

@inproceedings{liu2022convnet,
  title={A ConvNet for the 2020s},
  author={Liu, Zhuang and Mao, Hanzi and Wu, Chao-Yuan and Feichtenhofer, Christoph and Darrell, Trevor and Xie, Saining},
  booktitle={Proceedings of the IEEE/CVF Conference on Computer Vision and Pattern Recognition},
  pages={11976--11986},
  year={2022}
}

@inproceedings{
dosovitskiy2020image,
title={An Image is Worth 16x16 Words: Transformers for Image Recognition at Scale},
author={Alexey Dosovitskiy and Lucas Beyer and Alexander Kolesnikov and Dirk Weissenborn and Xiaohua Zhai and Thomas Unterthiner and Mostafa Dehghani and Matthias Minderer and Georg Heigold and Sylvain Gelly and Jakob Uszkoreit and Neil Houlsby},
booktitle={International Conference on Learning Representations},
year={2021},
url={https://openreview.net/forum?id=YicbFdNTTy}
}

@InProceedings{Liu_2021_ICCV,
    author    = {Liu, Ze and Lin, Yutong and Cao, Yue and Hu, Han and Wei, Yixuan and Zhang, Zheng and Lin, Stephen and Guo, Baining},
    title     = {Swin Transformer: Hierarchical Vision Transformer Using Shifted Windows},
    booktitle = {Proceedings of the IEEE/CVF International Conference on Computer Vision (ICCV)},
    month     = {October},
    year      = {2021},
    pages     = {10012-10022}
}

@inproceedings{wei2023fine,
  title={Fine-grained out-of-distribution detection of medical images using combination of feature uncertainty and mahalanobis distance},
  author={Wei, Jie and Wang, Guotai},
  booktitle={Proceedings of the IEEE International Symposium on Biomedical Imaging},
  pages={1--5},
  year={2023},
  organization={IEEE}
}

@inproceedings{anthony2023use,
  title={{On the use of Mahalanobis distance for out-of-distribution detection with neural networks for medical imaging}},
  author={Anthony, Harry and Kamnitsas, Konstantinos},
  booktitle={Proceedings of the International Workshop on Uncertainty for Safe Utilization of Machine Learning in Medical Imaging},
  year={2023},
  organization={Springer}
}

@article{hofmann2026ap,
  title={{AP-OOD: Attention Pooling for Out-of-Distribution Detection}},
  author={Hofmann, Claus and Huber, Christian and Lehner, Bernhard and Klotz, Daniel and Hochreiter, Sepp and Zellinger, Werner},
  journal={arXiv preprint arXiv:2602.06031},
  year={2026}
}

@inproceedings{hendrycks2019using,
  title={Using self-supervised learning can improve model robustness and uncertainty},
  author={Hendrycks, Dan and Mazeika, Mantas and Kadavath, Saurav and Song, Dawn},
  booktitle={Proceedings of the Advances in Neural Information Processing Systems},
  year={2019}
}

@inproceedings{hendrycks2022pixmix,
  title={{Pixmix: Dreamlike pictures comprehensively improve safety measures}},
  author={Hendrycks, Dan and Zou, Andy and Mazeika, Mantas and Tang, Leonard and Li, Bo and Song, Dawn and Steinhardt, Jacob},
  booktitle={Proceedings of the IEEE/CVF Conference on Computer Vision and Pattern Recognition},
  pages={16783--16792},
  year={2022}
}

@inproceedings{
tao2023non,
title={Non-parametric Outlier Synthesis},
author={Leitian Tao and Xuefeng Du and Jerry Zhu and Yixuan Li},
booktitle={The Eleventh International Conference on Learning Representations },
year={2023},
url={https://openreview.net/forum?id=JHklpEZqduQ}
}

@article{lu2024learning,
  title={Learning with mixture of prototypes for out-of-distribution detection},
  author={Lu, Haodong and Gong, Dong and Wang, Shuo and Xue, Jason and Yao, Lina and Moore, Kristen},
  journal={arXiv preprint arXiv:2402.02653},
  year={2024}
}

@article{ren2021simple,
  title={{A simple fix to Mahalanobis distance for improving near-OOD detection}},
  author={Ren, Jie and Fort, Stanislav and Liu, Jeremiah and Roy, Abhijit Guha and Padhy, Shreyas and Lakshminarayanan, Balaji},
  journal={arXiv preprint arXiv:2106.09022},
  year={2021}
}

@inproceedings{xmahalanobis2025,
  title = {{X-Mahalanobis: Transformer feature mixing for reliable OOD detection}},
  author = {Wei, Tong and Wang, Bo-Lin and Shi, Jiang-Xin and Li, Yu-Feng and Zhang, Min-Ling},
  booktitle = {Proceedings of the Annual Conference on Neural Information Processing Systems},
  year = {2025}
}

@inproceedings{mueller2025mahalanobisPP,
  title={{Mahalanobis++: Improving OOD detection via feature normalization}},
  author={M\"uller, Maximilian and Hein, Matthias},
  booktitle={Proceedings of the International Conference on Machine Learning},
  year={2025}
}

@inproceedings{liu2020energy,
  title={Energy-based out-of-distribution detection},
  author={Liu, Weitang and Wang, Xiaoyun and Owens, John and Li, Yixuan},
  booktitle={Proceedings of the Advances in Neural Information Processing Systems},
  year={2020}
}

@inproceedings{hendrycks2017baseline,
  title={A Baseline for Detecting Misclassified and Out-of-Distribution Examples in Neural Networks},
  author={Hendrycks, Dan and Gimpel, Kevin},
  booktitle={Proceedings of the International Conference on Learning Representations},
  year={2017}
}

@inproceedings{liang2018enhancing,
  title={Enhancing the Reliability of Out-of-distribution Image Detection in Neural Networks},
  author={Liang, Shiyu and Li, Yixuan and Srikant, Rayadurgam},
  booktitle={Proceedings of the International Conference on Learning Representations},
  year={2018}
}

@inproceedings{lee2018simple,
  title={A Simple Unified Framework for Detecting Out-of-Distribution Samples and Adversarial Attacks},
  author={Lee, Kimin and Lee, Kibok and Lee, Honglak and Shin, Jinwoo},
  booktitle={Proceedings of the Advances in Neural Information Processing Systems (NeurIPS)},
  volume={31},
  year={2018}
}

@inproceedings{sun2021react,
  title={{ReAct}: {O}ut-of-distribution Detection With Rectified Activations},
  author={Sun, Yiyou and Guo, Chuan and Li, Yixuan},
  booktitle={Proceedings of the Advances in Neural Information Processing Systems},
  volume={34},
  pages={144--157},
  year={2021}
}

@inproceedings{sun2022knn,
  title={Out-of-Distribution Detection with Deep Nearest Neighbors},
  author={Sun, Yiyou and Ming, Yifei and Zhu, Xiaojin and Li, Yixuan},
  booktitle={Proceedings of the International Conference on Machine Learning},
  pages={20827--20840},
  year={2022}
}

@inproceedings{liu2019large,
  title={Large-Scale Long-Tailed Recognition in an Open World},
  author={Liu, Ziwei and Miao, Zhongqi and Zhan, Xiaohang and Wang, Jiayuan and Gong, Boqing and Yu, Stella X},
  booktitle={Proceedings of the IEEE/CVF Conference on Computer Vision and Pattern Recognition (CVPR)},
  pages={2537--2546},
  year={2019}
}

@inproceedings{Rechtetal2019,
  author    = {Recht, Benjamin and Roelofs, Rebecca and Schmidt, Ludwig and Shankar, Vaishaal},
  title     = {Do ImageNet Classifiers Generalize to ImageNet?},
  booktitle = {International Conference on Machine Learning (ICML)},
  year      = {2019}
}

@article{Linetal2021,
  author  = {Lin, Yihan and Ding, Wei and Qiang, Shaohua and Deng, Lei and Li, Guoqi},
  title   = {ES-ImageNet: A Million Event-Stream Classification Dataset for Spiking Neural Networks},
  journal = {Frontiers in Neuroscience},
  year    = {2021},
  volume  = {15},
  doi     = {10.3389/fnins.2021.726582}
}

@inproceedings{
hendrycks2018benchmarking,
title={Benchmarking Neural Network Robustness to Common Corruptions and Perturbations},
author={Dan Hendrycks and Thomas Dietterich},
booktitle={International Conference on Learning Representations},
year={2019},
url={https://openreview.net/forum?id=HJz6tiCqYm},
}

@inproceedings{Hendrycksetal2021,
  author    = {Hendrycks, Dan and Basart, Steven and Mu, Norman and Kadavath, Saurav and Wang, Frank and Dorundo, Evan and Desai, Rahul and Zhu, Tyler and Samat, Samyak and Pimentel, Stephen and others},
  title     = {The Many Faces of Robustness: A Critical Analysis of Out-of-Distribution Generalization},
  booktitle = {Proceedings of the IEEE/CVF International Conference on Computer Vision (ICCV)},
  year      = {2021},
  pages     = {8340--8349}
}

@inproceedings{VanHornetal2018,
  author    = {Van Horn, Grant and Mac Aodha, Oisin and Song, Yang and Cui, Yin and Sun, Chen and Shepard, Alex and Adam, Hartwig and Perona, Pietro and Belongie, Serge},
  title     = {The iNaturalist Species Classification and Detection Dataset},
  booktitle = {2018 IEEE/CVF Conference on Computer Vision and Pattern Recognition (CVPR)},
  year      = {2018},
  pages     = {8769--8778},
  doi       = {10.1109/cvpr.2018.00914}
}

@article{Xiaoetal2013,
  author  = {Xiao, Jianxiong and Hays, James and Russell, Bryan C. and Patterson, Genevieve and Ehinger, Krista A. and Torralba, Antonio and Oliva, Aude},
  title   = {Basic level scene understanding: categories, attributes and structures},
  journal = {Frontiers in Psychology},
  year    = {2013},
  volume  = {4},
  doi     = {10.3389/fpsyg.2013.00506}
}

@article{fang2024eva,
  title={Eva-02: A visual representation for neon genesis},
  author={Fang, Yuxin and Sun, Quan and Wang, Xinggang and Huang, Tiejun and Wang, Xinlong and Cao, Yue},
  journal={Image and Vision Computing},
  volume={149},
  pages={105171},
  year={2024},
  publisher={Elsevier}
}
